\documentclass{article}
\usepackage{ijcai24}

\usepackage{xcolor}
\usepackage{mdframed}
\usepackage{times}
\usepackage{soul}
\usepackage{url}
\usepackage[hidelinks]{hyperref}
\usepackage[utf8]{inputenc}
\usepackage[small]{caption}
\usepackage{amsmath}
\usepackage{amsthm}
\usepackage{booktabs}
\usepackage{algorithm}
\usepackage{amssymb}
\usepackage[switch]{lineno}
\usepackage{wrapfig}
\usepackage{dsfont}
\usepackage[noend]{algpseudocode}
\usepackage{subfig} 
\usepackage{wrapfig}
\usepackage{graphicx}
\DeclareSubrefFormat{parens}{#1(#2)}
\newtheorem{Remark}{\indent Remark}
\newtheorem{Assumption}{\indent Assumption}

\title{vMFER: Von Mises-Fisher Experience Resampling\\ Based on Uncertainty of Gradient Directions for Policy Improvement }

\author{
Yiwen Zhu$^{1,2,3}$\and
Jinyi Liu$^{4}$\and
Wenya Wei$^1$\and
Qianyi Fu$^1$\and
Yujing Hu$^2$\footnotemark[1]\and
Zhou Fang$^1$\footnotemark[1]\and \\
Bo An$^{3,5}$\and
Jianye Hao$^4$\and
Tangjie Lv$^2$\And
Changjie Fan$^2$\\
\affiliations
$^1$Zhejiang University \\
$^2$NetEase Fuxi AI Lab \\
$^3$Nanyang Technological University\\
$^4$Tianjin University \\
$^5$Skywork AI \\
\emails
\{evanzhu, wwy\_vivian, qyfu, zfang\}@zju.edu.cn,\;
\{jyliu, jianye.hao\}@tju.edu.cn,\\
\{huyujing, hzlvtangjie, fanchangjie\}@corp.netease.com,\;
boan@ntu.edu.sg\;
}

\begin{document}

\maketitle

\begin{abstract}
    Reinforcement Learning (RL) is a widely employed technique in decision-making problems, encompassing two fundamental operations -- policy evaluation and policy improvement. Enhancing learning efficiency remains a key challenge in RL, with many efforts focused on using ensemble critics to boost policy evaluation efficiency.
    However, when using multiple critics, the actor in the policy improvement process can obtain different gradients. Previous studies have combined these gradients without considering their disagreements. Therefore, optimizing the policy improvement process is crucial to enhance learning efficiency. This study focuses on investigating the impact of gradient disagreements caused by ensemble critics on policy improvement. We introduce the concept of uncertainty of gradient directions as a means to measure the disagreement among gradients utilized in the policy improvement process. Through measuring the disagreement among gradients, we find that transitions with lower uncertainty of gradient directions are more reliable in the policy improvement process. Building on this analysis, we propose a method called von Mises-Fisher Experience Resampling (vMFER), which optimizes the policy improvement process by resampling transitions and assigning higher confidence to transitions with lower uncertainty of gradient directions. Our experiments demonstrate that vMFER significantly outperforms the benchmark and is particularly well-suited for ensemble structures in RL.
\end{abstract}

\section{Introduction}
\footnotetext[1]{The corresponding authors.}

Over the past few years, there has been rapid progress in the field of reinforcement learning (RL), leading to impressive achievements in tackling complex tasks \cite{wu2023daydreamer,radosavovic2023real,abeyruwan2023sim2real}. Despite these advancements, the challenge of enhancing learning efficiency persists.

In general, reinforcement learning involves two fundamental operations: policy evaluation and policy improvement \cite{sutton2018reinforcement}. 
To enhance learning efficiency and optimality, numerous methods optimize the policy evaluation process by using ensemble critics, such as Double Q-learning \cite{hasselt2010double}, SAC \cite{haarnoja2018soft,haarnoja2018soft2}, TD3 \cite{fujimoto2018addressing} and REDQ \cite{chen2021randomized}.
Nevertheless, the utilization of ensemble critics often introduces disagreements in the direction of policy optimization during the policy improvement process.
Existing methods like SAC, TD3, or REDQ simply aggregate the multiple gradients generated during policy improvement into a single gradient, without considering the disagreements among these gradients caused by ensemble critics.
One alternative approach is to enhance the reliability of gradients in policy improvement, such as utilizing the delayed policy update method employed by REDQ and TD3.
This method uses more reliable ensemble critics to ensure a more concentrated gradient direction.
However, this approach does not account for the discrepancies among transitions, resulting in delayed updates for all sampled transitions.

We propose that by selectively avoiding delayed updates for transitions that can provide a reliable gradient under the current ensemble critics, the policy improvement process can be further optimized.
We introduce additional indicators to measure the reliability of Q-ensembles under current ensemble critics.
This allows us to identify which transition data is more appropriate for utilization in the policy improvement process.
We posit that as the accuracy of the ensemble critics increases, the directions of policy gradients provided by the same transition under the ensemble structure will demonstrate a high concentration in the policy improvement process.
Hence, we introduce the concept of uncertainty of gradient directions to identify the reliability of transitions under the current ensemble structure during the policy improvement process.
From a directional statistics perspective \cite{mardia2000directional}, these directions of the policy gradients can be modeled as a distribution.
Considering the computational cost, we use the von Mises-Fisher distribution \cite{fisher1953dispersion} to quantify such uncertainty associated with each transition.
Furthermore, we propose the von Mises-Fisher Experience Resampling (vMFER) algorithm which leverages the uncertainty of gradient directions to resample transitions for policy improvement.
To improve the efficiency of the policy improvement process, we enhance the sampling probability of transitions with lower uncertainty while reducing the likelihood of sampling transitions with higher uncertainty during the policy improvement process.

Our primary contributions are threefold: 
\begin{enumerate}%[1)]
    \item We introduce a metric to measure the uncertainty of gradient directions, aimed at evaluating the reliability of transitions used in the policy improvement process. This metric is calculated by analyzing the discrepancy in gradient directions, which are induced by ensemble critics for each transition.     
    \item We propose the vMFER algorithm to optimize policy improvement by resampling transitions based on the uncertainty of gradient directions. Moreover, it is compatible with most actor-critic algorithms utilizing the ensemble structure.
    \item Our approach performs effectively in Mujoco control tasks \cite{brockman2016openai}. This indicates the potential of vMFER for a wide range of applications.
\end{enumerate}

\section{Preliminary}
\subsection{Actor-critic Framework}

The Actor-Critic framework is widely used in RL, consisting of two distinct modules: the actor network that learns the policy, and the critic network that learns the value function.

Several RL algorithms have been proposed based on the actor-critic framework. Algorithms like PPO \cite{schulman2017proximal} and DDPG \cite{lillicrap2015continuous} use a single critic structure, while others like TD3 \cite{fujimoto2018addressing} and SAC \cite{haarnoja2018soft} use an ensemble structure with multiple critics to overcome the problem of overestimation, arising due to the maximization of a noisy value estimate during the critic learning process \cite{thrun1993issues}.
The ensemble structure results in multiple Q-values for each transition, allowing for the calculation of multiple actor network losses and generating multiple gradients for actor network parameter updates. 
The loss function for a mini-batch of transitions in actor training is commonly expressed as $\mathds{E}_{(s,a)\sim D} \left[\log\pi(a|s) - \mathop{\min}\limits_i {Q_i}(s,a)\right]$ \cite{haarnoja2018soft}, $\mathds{E}_{(s,a)\sim D} \left[-Q_1(s,a) \right]$ \cite{lillicrap2015continuous,fujimoto2018addressing}, or $ \mathds{E}_{(s,a)\sim D} \left[ \frac{1}{N}\sum\limits_i^N \left[\log\pi - Q_i(s,a)\right]\right]$ \cite{chen2021randomized}. Here, $D$ represents the replay buffer, and the subscript of critic $Q$ denotes the index number of ensemble critics, $\pi$ refers to the policy.

The gradients of mini-batch transitions provided for policy updates are usually integrated by averaging these gradients to update actor network parameters.

\subsection{Von Mises-Fisher Distribution}\label{SE:p_vmfdis}

The von Mises-Fisher (vMF) distribution \cite{fisher1953dispersion} is one of the most basic probability distributions in high-dimensional directional statistics \cite{mardia2000directional}. 
It characterizes a probability distribution on the {$(p-1)$}-sphere in $\mathds{R}^{p}$, defined on the unit hypersphere.
To be more specific, the probability density function of the von Mises-Fisher distribution for a random $p$-dimensional unit vector $\mathbf{x} \sim \text{vMF}(k,\mu)$ is expressed in Eq. ({\ref{Eq：vmf_pdf}}), where $f_p$ represents the density function for $ \text{vMF}(k,\mu)$.
\begin{align}
\begin{split}
\mathbf{x}&\sim \text{vMF}(k,\mu), \\
    f_p(\mathbf{x} ;\mathbf{\mu},k) &= C_p(k)\exp(k\mathbf{\mu}^\text{T}\mathbf{x}).
\end{split}
\label{Eq：vmf_pdf}
\end{align}
In this function, $C_p(k)$ represents the normalization constant, 
and $k \geq 0$  represents the concentration parameter. Furthermore, the mean direction $\mathbf{\mu}$ can be calculated as demonstrated in Eq. ({\ref{Eq:meandir_vmf}}) \cite{mardia2000directional}.
\begin{align}
    \mathbf{\mu}={{\Bar{\mathbf{x}}}}/{{\Bar{\mathbf{R}}}}, \text{where}\  \Bar{\mathbf{x}}=\frac{1}{N}\sum\limits_i^N x_i,\ \Bar{\mathbf{R}}=||\Bar{\mathbf{x}}||_2 .
    \label{Eq:meandir_vmf}
\end{align}

The concentration parameter $k$ is commonly used to indicate the degree of clustering and scattering of the vector direction distribution.
However, estimating the concentration parameter $k$ using the Maximum-likelihood estimate is often challenging because of the difficulty in calculating the modified Bessel function \cite{sra2012short}. 
Banerjee et al. \cite{banerjee2005clustering} proposed a simpler approximation given by: 
\begin{align}
    \hat{k} = \frac{\Bar{\mathbf{R}}(p-\Bar{\mathbf{R}}^2)}{(1-\Bar{\mathbf{R}}^2)}.
    \label{Eq:k}
\end{align}
which avoids calculating Bessel functions \cite{sra2012short}.

\section{Von Mises-Fisher Experience Resampling}

In the actor-critic framework with the ensemble structure, each ensemble critic can theoretically contribute a gradient for updating the actor during the policy improvement process. 
Despite this, existing Actor-Critic frameworks  \cite{fujimoto2018addressing,haarnoja2018soft,lillicrap2015continuous,chen2021randomized} often merge these multiple gradients into a single gradient for policy improvement, disregarding the information that can be derived from the discrepancies among these gradients.
Specifically, these frameworks often overlook the uncertain information in different gradients generated by the same transition.
This information can be utilized to evaluate the reliability of the gradient provided by that particular transition for policy improvement.

\begin{figure}[!t]
\vspace{0pt}
% \centering
\includegraphics[width=0.99\columnwidth]{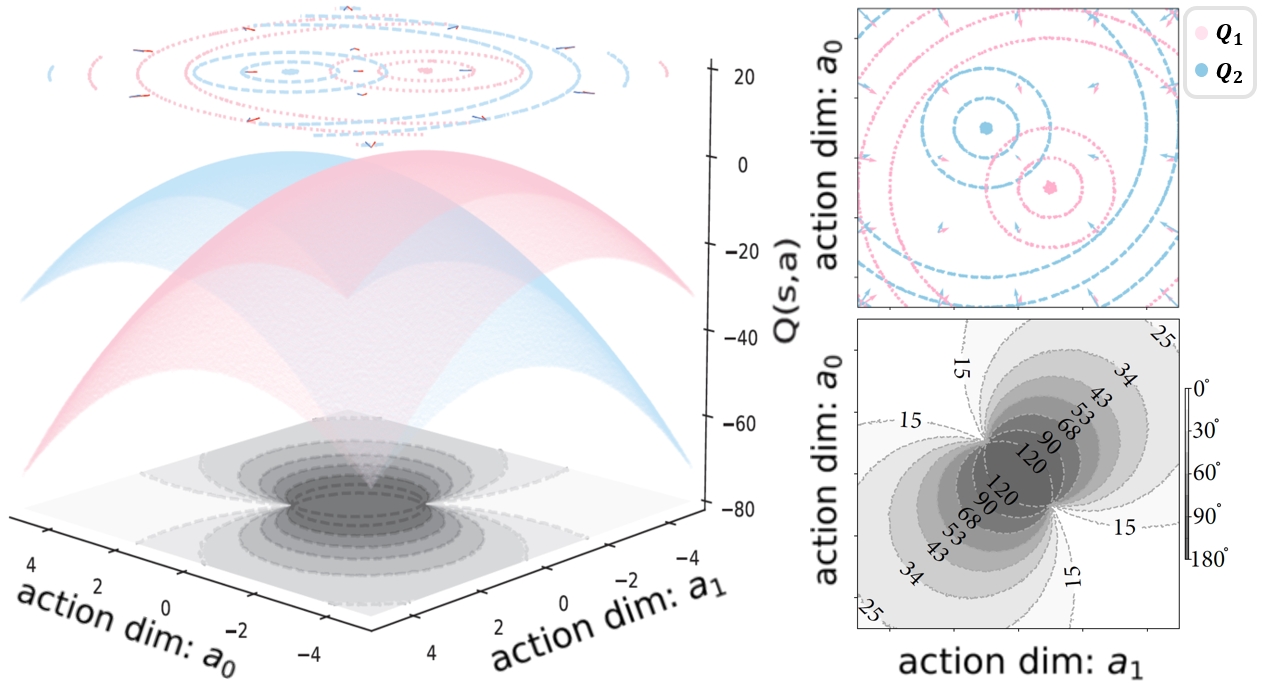}
\label{Fig:Q1Q2}
\caption{  Multiple Q-values and their corresponding gradients $\frac{\partial Q_i(s_t,a)}{\partial a}$ are generated by the ensemble critic function for a given state input $s_t$, with each Q-value and gradient pair corresponding to a different action $a$. \textbf{Left:} The ensemble critic function $Q_1(s_t,\cdot)$ and  $Q_2(s_t,\cdot)$, which are depicted as multi-dimensional surfaces. \textbf{Right Upper:}The gradients of the ensemble critics, represented as arrows with varied colors on the contour plots of the Q-values, illustrate the direction and magnitude of the action-value function's sensitivity to changes in action space. \textbf{Right Lower:} The angles between the gradients, $\frac{\partial Q_1(s_t,a)}{\partial a}$ and  $\frac{\partial Q_2(s_t,a)}{\partial a}$, are used to quantify the uncertainty of gradients for different actions $a$. }
\label{Fig:uncertaintyOfGradients}
\end{figure}

To address these limitations, this paper proposes the use of the von Mises-Fisher distribution to describe the uncertainty of gradient directions. 
Subsequently, we propose the von Mises-Fisher Experience Resampling (vMFER) algorithm,
which involves resampling the transitions using probabilities calculated based on the uncertainty of gradient directions for policy improvement.
By decreasing the probability of sampling transitions with high uncertainty, the policy improvement can be made more reliable.
Furthermore, we provide a straightforward example and a toy experiment to demonstrate the uncertainty of gradient directions in the ensemble structure, before explaining in-depth how this indicator contributes to enhancing algorithm performance.

\subsection{
Exploring Disagreements in Gradient Directions: A Simple Example
}\label{SE:Asimpleexample}

To interpret the uncertainty of gradient directions, we present an example of policy evaluation on a two-critic ensemble.
To highlight the disagreement among ensemble critics during the learning process, we have formulated two critics given current state $\mathbf{s_t}$ and action $\mathbf{a}$: 
\begin{align}
    \begin{split}
        Q_1(\mathbf{s_t},\mathbf{a}) &= -(\mathbf{a}+\mathds{1}_{2\times1}+\epsilon)^T(\mathbf{a}+\mathds{1}_{2\times1}+\epsilon),\\
    Q_2(\mathbf{s_t},\mathbf{a}) &= -(\mathbf{a}-\mathds{1}_{2\times1}+\epsilon)^T(\mathbf{a}-\mathds{1}_{2\times1}+\epsilon).
    \end{split}
\end{align}
where $\mathbf{a}\in\mathds{R}^{{2\times1}}$ and $\epsilon\sim N(0,0.01)$. 
The left of Figure \ref{Fig:uncertaintyOfGradients} shows the variations of the output $Q$ for different action inputs of the two critic networks. The three axes represent two dimensions of the action and the value of $Q$, respectively.

Then we establish the optimization objective for the actor network, similar to the conventional continuous RL: {\scriptsize $ \mathop{max}\limits_{\mathbf{a}} Q(\mathbf{s_t},\mathbf{a})$}. This allows us to compute the gradient {\footnotesize ${ \frac{\partial Q(\mathbf{s_t},\mathbf{a})}{\partial \mathbf{a}}}$} and determine the convergence direction for the desired action based on the current critic network and $s_t$.
Due to the ensemble critics, multiple gradients can be computed for each transition in the policy improvement process, as shown in the right upper of Figure \ref{Fig:uncertaintyOfGradients}.
From the perspective of policy improvement, 
the presence of disagreements among gradients becomes apparent, underscoring the importance of their judicious utilization.

Obviously, a metric is required to quantify the disagreement in gradients during the policy improvement process.
In this study, we employ the uncertainty of the gradient directions as a measure of the extent of these disagreements. 
As demonstrated in the right lower of  Figure \ref{Fig:uncertaintyOfGradients}, the larger angle represents more significant disagreements among gradient directions, indicating substantial conflicts in gradients for specific action inputs.
Using this metric, we can assign a confidence level to the corresponding transition, where the magnitude of confidence should be inversely proportional to the level of uncertainty.

\subsection{The Necessity of Measuring Gradient Uncertainty: A Toy Experiment}\label{SE:NecOFun}
\begin{figure*}[!htbp]
\centering
    \subfloat[Shooting environment: an one step MDP]
    {
    \includegraphics[width=.45\columnwidth,height = 2.7 cm]{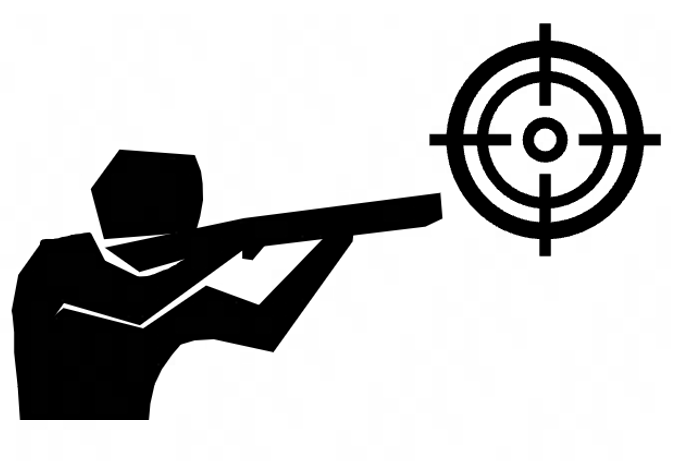}
    \label{Fig:toyenv}
    }
\hspace{5pt}
\subfloat[ The distributions of angles between gradients during training process by various approaches]
 {
 \includegraphics[width=.75\columnwidth,height = 3.0cm]{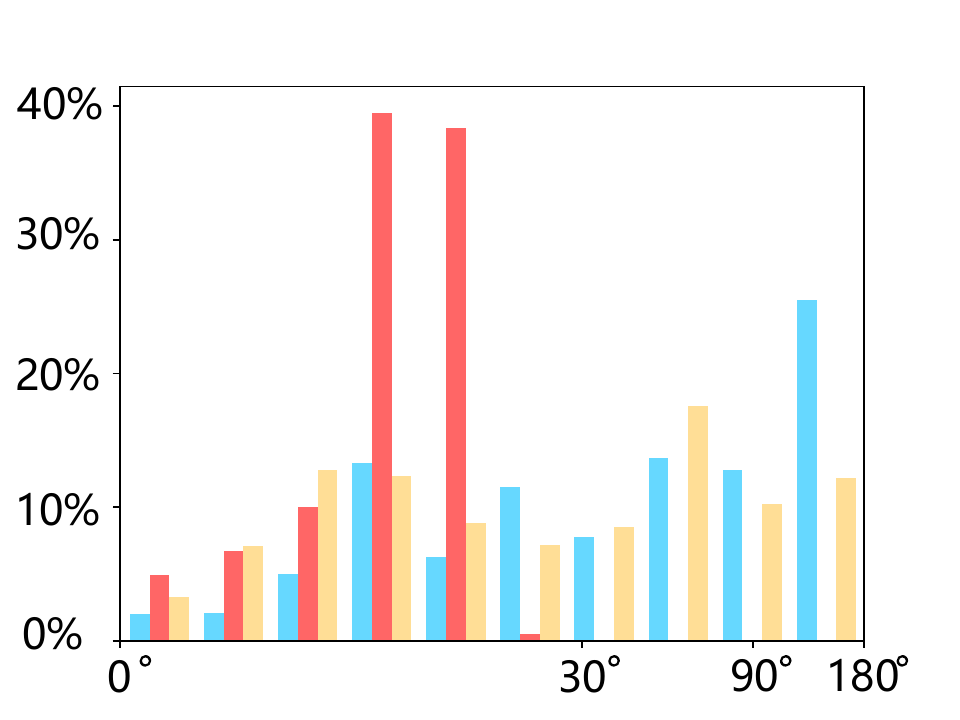}
  \label{Fig:toyhis}
 }
 \hspace{5pt}%TheUncertaintyofGradients_11961
\subfloat[ Episode rewards achieved by various approaches]
 {
 \includegraphics[width=.6\columnwidth,height = 3.0cm]{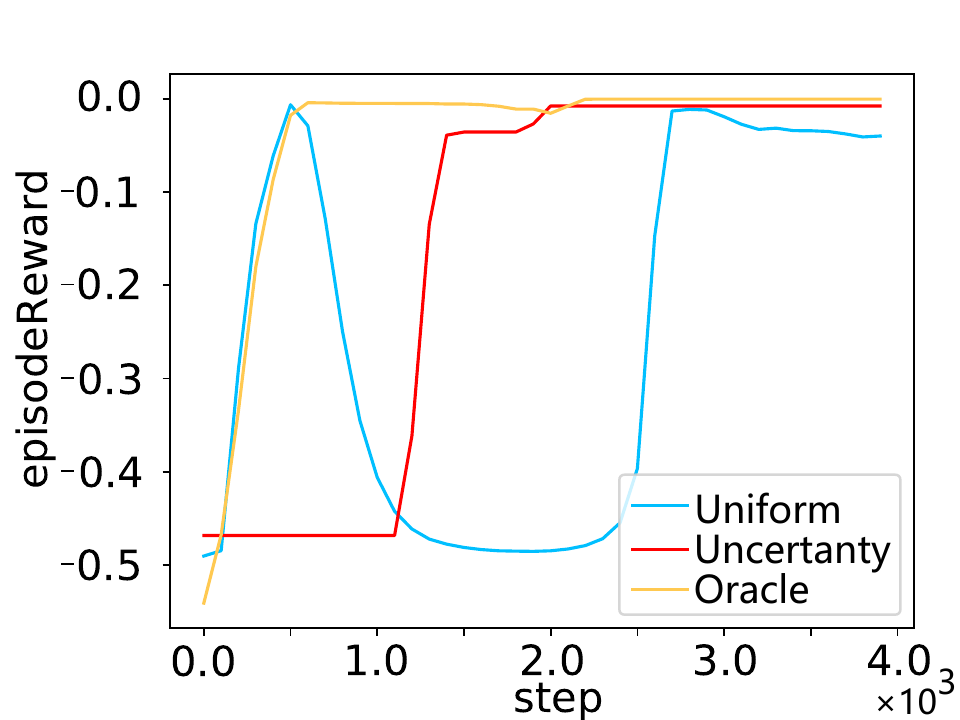}
  \label{Fig:toyepr}
 }
 
 \subfloat[ Trajectory of policy changes during update]
 {
 \includegraphics[width=2\columnwidth]{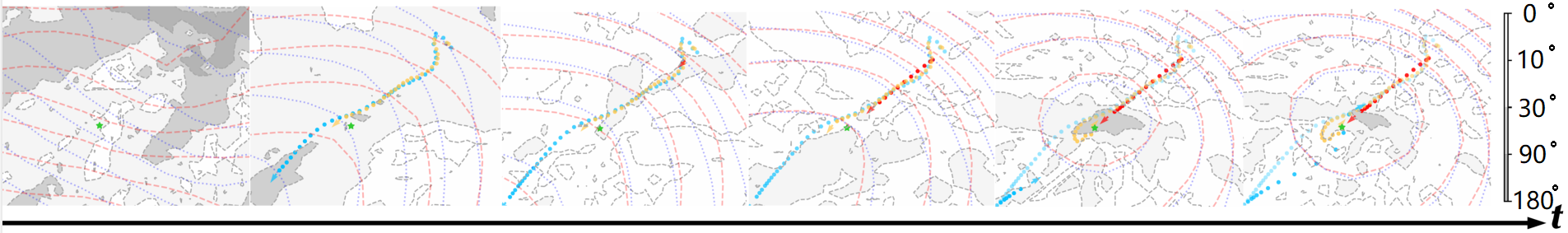}
  \label{Fig:toytraj}
 }
 
\caption{A toy experiment illustrating the advantage of considering the uncertainty of gradient directions on the learning efficiency of policy improvement. The experiment compares three approaches: `\textbf{Uniform}' involves uniformly sampling from the transitions, `\textbf{Uncertainty}' utilizes only transitions with low uncertainty of gradient directions, and `\textbf{Oracle}' employs only transitions that update the action in the direction of the optimal action.}
\label{Fig:toyexp}
\end{figure*}

To demonstrate the influence of resampling on policy improvement, we introduce an artificial environment called the ``Shooting'' environment (depicted in Figure \ref{Fig:toyexp}\subref{Fig:toyenv}). 
This environment is a one-step Markov Decision Process (MDP) with a continuous action space, and the optimal action is indicated by a green star in Figure \ref{Fig:toyexp}\subref{Fig:toytraj}. 
The closer the policy's action output is to the optimal action, the higher the reward obtained.

In this experimental setup, we employ three distinct approaches to investigate the effects of resampling transitions during the policy improvement phase in such a one-step MDP environment. The `\textbf{Uniform}' method involves uniformly sampling transitions. In contrast, the `\textbf{Uncertainty}' method selectively uses transitions with lower uncertainty of gradient directions. Finally, the `\textbf{Oracle}' method chooses transitions that can guide the updated action toward the optimal outcome.

As we focus on a one-step MDP, the actions generated by the actor exhibit variation when subjected to different policy improvement techniques during training, yet the policy evaluation remains constant. 
This dynamic is carefully traced and depicted in Figure \ref{Fig:toyexp}\subref{Fig:toytraj}.
Consistent with the setting outlined in Section~\ref{SE:Asimpleexample}, we utilize red and blue contour maps to represent Q-values.
The uncertainty of gradient directions is quantified by measuring the angles between these gradients.
Furthermore, this uncertainty is visually captured through a grey contour map, where darker shades indicate a higher degree of discrepancy in gradient directions for specific action inputs, as depicted in Figure \ref{Fig:toyexp}\subref{Fig:toytraj}.
Figure \ref{Fig:toyexp}\subref{Fig:toyepr} presents a comparison of episode rewards achieved by agents trained using different approaches.
Additionally, Figure \ref{Fig:toyexp}\subref{Fig:toyhis} captures the distribution of angles between gradients associated with the actions output by policies, which have been updated through various methods across the entire training process. Notably, we find that the `\textbf{Oracle}' and `\textbf{Uniform}' methods yield a relatively even distribution of uncertainty levels in the transitions used during the policy improvement phase. Conversely, the `\textbf{Uncertainty}' method tends to select transitions with lower uncertainty for policy improvement, namely smaller angles.

In contrast, the `\textbf{Oracle}' method represents an ideal approach. However, its practical application is limited due to the prerequisite of knowing the optimal policy beforehand, which is often not feasible in real-world scenarios. Intriguingly, while the `\textbf{Uncertainty}' method may exhibit lower efficiency in policy improvement compared to the `\textbf{Oracle}' method, it excels in terms of the policy update trajectory and offers greater practical applicability. Moreover, both the `\textbf{Oracle}' and `\textbf{Uncertainty}'  outperform the `\textbf{Uniform}' approach.

Our research indicates that the careful selection of transitions for policy improvement is crucial for enhancing learning efficiency. 
Specifically, the resampling method that leverages the uncertainty of gradient directions is effective in optimizing the learning process.
Essentially, less concentrated gradient directions signal higher uncertainty, with more uncertainty indicating greater divergence in these directions.
\begin{Remark}
Transitions under the current ensemble critics with higher uncertainty of gradient directions should be less likely to be employed for policy improvement.
\end{Remark}

\subsection{How To Measure Gradient Uncertainty: Via Von Mises-Fisher Distribution}\label{SE:howtomeasure}

Our aim is to identify a metric that can determine which transitions contribute reliable gradients for actor updates.
Figure {\ref{Fig:uncertaintyOfGradients}} illustrates that this metric should describe the concentration of gradients contributed by the same transition under different indices of the ensemble critics.
It's essential to emphasize that, in this context, the direction of the gradient is more crucial than its length. 
Comparatively, an incorrect gradient descent direction is less acceptable than an incorrect magnitude. Because, unlike the latter, the former implies ineffective optimization.
Moreover, the metric is not too complex to compute, since in theory, we need to compute the metric for each transition before updating the actor.

In the field of directional statistics, few distributions align with our criteria. The Bingham \cite{bingham1974antipodally} and Kent \cite{kent1982fisher} distributions, while noteworthy, require the computation of the Bessel function \cite{bowman2012introduction}, thereby not fulfilling our need for low computational overhead. 
In contrast, the von Mises-Fisher distribution \cite{fisher1953dispersion,watson1982distributions,mardia2000directional}, particularly when employing Banerjee's \cite{banerjee2005clustering} method for parameter estimation, circumvents the need for Bessel function calculations. 
Notably, the concentration parameter $k$ we required is expressed simply. 
The efficacy and accuracy of this approximation method are well-demonstrated by \cite{sra2012short}.
Additionally, employing vMF to model gradient directions offers advantages of scalability and a threshold-free setup, compared to using the angle between gradients as mentioned in Section~\ref{SE:Asimpleexample} and Section~\ref{SE:NecOFun}.

Hence, we assume that the directions of the gradients are sampled from the von Mises-Fisher distribution.
To measure the uncertainty of the gradient directions, the parameters of the distribution need to be estimated.
Using the ensemble structure enables us to compute numerous actor losses and their corresponding gradients with respect to the actor network parameters $\theta$ using the same transition $(s_t,a_t,r_t,s'_t)$, as illustrated in Eq. ({\ref{Eq:losses}}).
\begin{align}
\small
\begin{split}
\mathbf{L}{(s_t,a)} &=\left [ \begin{matrix}
l_{1}(s_t,a)&\cdot\cdot\cdot &l_{n}(s_t,a) 
\end{matrix} \right ]^{\text{T}},\\
\frac{\partial \mathbf{L}(s_t,a)}{\partial \theta} &=  \left [ 
\begin{matrix}
\frac{\partial l_{1}(s_t,a)}{\partial \theta }  &\cdot\cdot\cdot &  \frac{\partial l_{n}(s_t,a)}{\partial \theta } 
\end{matrix} 
\right ]^{\text{T}}, \; a\sim\pi(\cdot|s_t).\\
\end{split}
\label{Eq:losses}
\end{align}
where $n$ is the ensemble size of the critic networks.
Estimating the parameter of the Von Mises-Fisher distribution for $\frac{\partial l_i}{\partial \theta}$ can be challenging due to the high dimensionality of $\theta$. This can result in issues such as increased computational cost and a large scale of the estimated concentration parameter $k$.
To mitigate these issues, it is advisable to reduce the dimensionality of the gradients. 
By applying the chain rule, it is evident that $\frac{\partial a}{\partial \theta}$ is constant for the same transition. 
As a result, the uncertainty in the gradient directions primarily arises from $\frac{\partial l_i}{\partial a}, i\in[1,n]$. Therefore, calculating the uncertainty of $\frac{\partial l_i}{\partial a}$ rather than $\frac{\partial l_i}{\partial \theta}$ is a more cost-effective and scalable approach.
\begin{align}
    \begin{split}
         \frac{\partial l_i(s_t,a )}{\partial \theta} 
         = \frac{\partial l_i(s_t,a )}{\partial a} \frac{\partial a}{\partial \theta}, \qquad a\sim \pi(\cdot|s_t).
    \end{split}
    \label{Eq:chainrule}
\end{align}

Let {\footnotesize$x_i(s_t) = ||\frac{\partial l_i(s_t,a)}{\partial a}||_2^{-1} \boldsymbol{\cdot} \frac{\partial l_i(s_t,a)}{\partial a} |_{a\sim\pi(\cdot|s_t)}\; $} and $\;\mathbf{x}(s_t) =\sum_i x_i(s_t) /{n}$, where $x_i(s_t)$ denote the direction of the gradient contributed by the current actor loss, assuming that $x_i \sim \text{vMF}(k,\mu)$, then according to Banerjee's method we can estimate the concentration parameter $k$ and mean direction $\mu$ as demonstrated in Eq. (\ref{Eq:rmuk}).
\begin{align}
\begin{split}
        \mathbf{R}(s_t) &= ||\mathbf{x}(s_t) ||_2, \quad \mu(s_t)  = \frac{ \mathbf{x}(s_t)}{\mathbf{R}(s_t)},\\
        k(s_t) &= \frac{\mathbf{\mathbf{R}(s_t)}(p-\mathbf{R}^2(s_t))}{(1-\mathbf{R}^2(s_t))} \quad\propto \quad\mathbf{R}(s_t).
\end{split}
\label{Eq:rmuk}
\end{align}
Here, $p$ denotes the dimension of the action. The concentration parameter $k$ is used to articulate the uncertainty present in the current gradient directions. 
 Clearly, we can prove that $\mathbf{R}$ is proportional to $k$.  Hence, $\mathbf{R}$ possesses a similar capability to represent uncertainty.

\begin{algorithm}[!t]
\caption{Von Mises-Fisher Experience Resampling (Based on TD3)}
\label{Algo:vMFER_td3}
\footnotesize 
\begin{algorithmic}[1]
\State
    Initialize replay buffer $\mathcal{D}$ and ensemble number $N$
\State  
    Initialize critic networks $\{Q_{\phi_i}\;|\; i\in[1,N]\}$, and actor network $\pi_\theta$ with random parameters $\{{\phi_i}\;|\; i\in[1,N]\}$, $\theta$
\State
    Initialize target network $\{{\phi_i}{'} \leftarrow {\phi_i}\;|\; i\in[1,N]\}$, $\theta{'} \leftarrow \theta$

\State
    Initialize sampling factors $p_j = 1$ for each transition of $\mathcal{D}$
\For{$t$=1 to $T$ }
    \State sample action with noise $a\leftarrow\pi_{\theta}(s)+\epsilon,\; \epsilon \sim \mathcal{N}(0,\sigma)$
    \State store transition $(s,a,r,s')$ in $\mathcal{D}$ 
    \State sample mini-batch of $b$ transitions $(s,a,r,s')$ from $\mathcal{D}$
    \State $a'\leftarrow \pi_\theta(s') + \epsilon,\; \epsilon \sim clip( \mathcal{N}(0,\sigma),-c,c)$
    \State $y\leftarrow r+ \gamma \min_{i=1,2}Q_{\phi_i}(s',a')$
    \State Update critics $\phi_i \leftarrow \mathop{argmin}_{\phi_i} b^{-1}\sum (y - Q_{\phi_i}(s,a))^2$
    % \State update critics as in TD3 (only use the )
    \If{ $t$ mod 2}
    \For{$j = 1$ to $b$}
    \State {\color{brown} \textbackslash\textbackslash Resample transition}
    \State  $(s_j,a_j,r_j,s'_j)\sim P(j) = \frac{p_j}{\sum_m p_m}$ \Comment{ Eq. ({\ref{Eq:Baye}})}
    \State \textbackslash\textbackslash Sample action 
    \State$\hat{a}_j\leftarrow \pi_{\theta}(s_j) + \epsilon, \;\epsilon \sim clip(\mathcal{N}(0,\sigma),-c,c)$
    \State {\color{brown} \textbackslash\textbackslash Calculate the actor losses}
     \State   $l_i(s_j,\hat{a}_j) = -Q_{\phi_{i} }(s_j,\hat{a}_j),\;  \;i\in [1,N] $ 
    \State {\color{brown} \textbackslash\textbackslash Calculate gradients of losses}
    \State $ g_i = \frac{\partial l_i(s_j,a)}{\partial a} |_{a=\hat{a}_j},\;  \;i\in [1,N] $ 
    \State $p_j\leftarrow$ {\color{brown} Update Sampling Factor $p_j$ (Algorithm \ref{Algo:usfpj})}
    \EndFor
    \State Update $\theta$ by the deterministic policy gradient:
    \State $\nabla_\theta J(\theta) =-b^{-1}\sum_j \nabla_\theta l_1(s_j)$
    \State Update target networks:
    \State$\phi_i'\leftarrow \tau \phi_i+(1-\tau)\phi_i'$, $\theta'\leftarrow\tau\theta+(1-\tau)\theta'$
    \EndIf
\EndFor
\end{algorithmic}
\end{algorithm}

\subsection{How To Use vMF Distribution: Von Mises-Fisher Experience Resampling}\label{SE:Howtousevmf}

Our objective is to independently fit von Mises-Fisher distributions to the gradients from each transition and evaluate their uncertainty levels to ascertain the probability of each transition's utilization. A higher level of uncertainty implies a reduced likelihood of sampling the data. The gradient directions of each data corresponds to a distinct von Mises-Fisher distribution.
We define the likelihood of the prior distribution for each transition being sampled as {\footnotesize $P(j|\mathcal{D}) = \frac{1}{M}$}, where $\mathcal{D}$ denotes the transitions in the replay buffer and $M$ the total number of data points in it.
Subsequently, we represent the conditional probability distribution as {\footnotesize $P(x(s_j)| j,\mathcal{D})= C_p(k(s_j))\exp(k(s_j)\mu^{\text{T}}(s_j)x(s_j))$}. The posterior distribution, which we aim to achieve, is denoted by {\footnotesize $P(j| x(s_j),\mathcal{D}) = \frac{p_j}{\sum_m p_m}$}. 
Here, $p_j$ indicates the sampling factor for the specific transition $(s_j,a_j,r_j,s'_j)$.
This framework enables us to derive the posterior distribution of the resampling probability for the current transition after sampling the gradient direction $x(s_j) \sim \text{vMF}(k(s_j),\mu(s_j))$, as elaborated in Eq. ({\ref{Eq:Baye}}).
\begin{align}
\begin{split}
        &\prod \limits^M_{ j=1} \frac{1}{M} C_p(k(s_j))\exp({k(s_j)\mu(s_j)^\text{T}x(s_j)}) 
    \\\propto &
  \prod\limits_{j=1}^{M} P(j|x(s_j),\mathcal{D}) 
  =\prod\limits_{j=1}^{M}\frac{p_j}{\sum_m p_m} .
\end{split}
  \label{Eq:Baye}
\end{align}
As $k\propto R$, and the dimensionality of action heavily affects the value range of $k$, we simplify the calculation and set the probability of sampling each transition using  Eq. ({\ref{Eq:prob}}).
\begin{align}
    P(j|x(s_j),\mathcal{D}) = \frac{\exp(\mathbf{R}(s_j)\mu^{\text{T}}(s_j) x(s_j))}{\sum\limits_i^M \exp(\mathbf{R}(s_i)\mu^{\text{T}}(s_i) x(s_i))}.
    \label{Eq:prob}
\end{align}

\begin{algorithm}[!t]
\caption{Update Sampling Factors $p_j$ }
\renewcommand{\algorithmicrequire}{\textbf{Input:}}
\renewcommand{\algorithmicensure}{\textbf{Output:}}
\label{Algo:usfpj}
\footnotesize
\begin{algorithmic}[1]
\Require $p_j$, $s_j$, $\{g_1,g_2,\cdot\cdot\cdot,g_N\}$, $\{Q_{\phi_1},Q_{\phi_2},\cdot\cdot\cdot,Q_{\phi_N}\}$,$\pi_\theta$
\State Calculate normalized unit vector $x_i = \frac{g_i}{||g_i||_2}$, $i \in [1,N]$
\State Calculate other parameters:\\
\qquad$\mathbf{R}(s_j) = ||\frac{\sum_i x_i }{N}||_2$ and $\mu(s_j) = \frac{\sum_i x_i }{N \mathbf{R}(s_j) }$ \Comment{Eq. (\ref{Eq:rmuk})}
\State Choose index of critics used in the policy improvement: \\
% Achieve the index with different methods 
    \qquad$ e = \left\{ {\scriptsize 
    \begin{aligned}
    & \mathop{argmin}\limits_i Q_{\phi_i}(s_j,\pi_\theta(s_j))  \quad& (\textbf{SAC}) \\
    & 1 & (\textbf{TD3})
    \end{aligned} }
\right. $
\State  $p_j\leftarrow  
\left\{ {\scriptsize
    \begin{aligned}
    & \exp(\mathbf{R}(s_j)\mu^{\text{T}}(s_j)x_{e}(s_j)) \;\quad(\textbf{uncertainty})  \\
    & {\footnotesize rank(\exp(\mathbf{R}(s_j)\mu^{\text{T}}(s_j)x_{e}(s_j)))^{{-1}} }(\textbf{rank})
    \end{aligned} }
\right. $ \Comment{ Eq. ({{\ref{Eq:prob},\ref{Eq:rankprob}}})}
\Ensure $p_j$
\end{algorithmic}
\end{algorithm}

\begin{figure*}[!htbp]
\centering

    \hspace{-6mm}
    \subfloat[Hopper-TD3]{\includegraphics[width=.38\columnwidth  ]{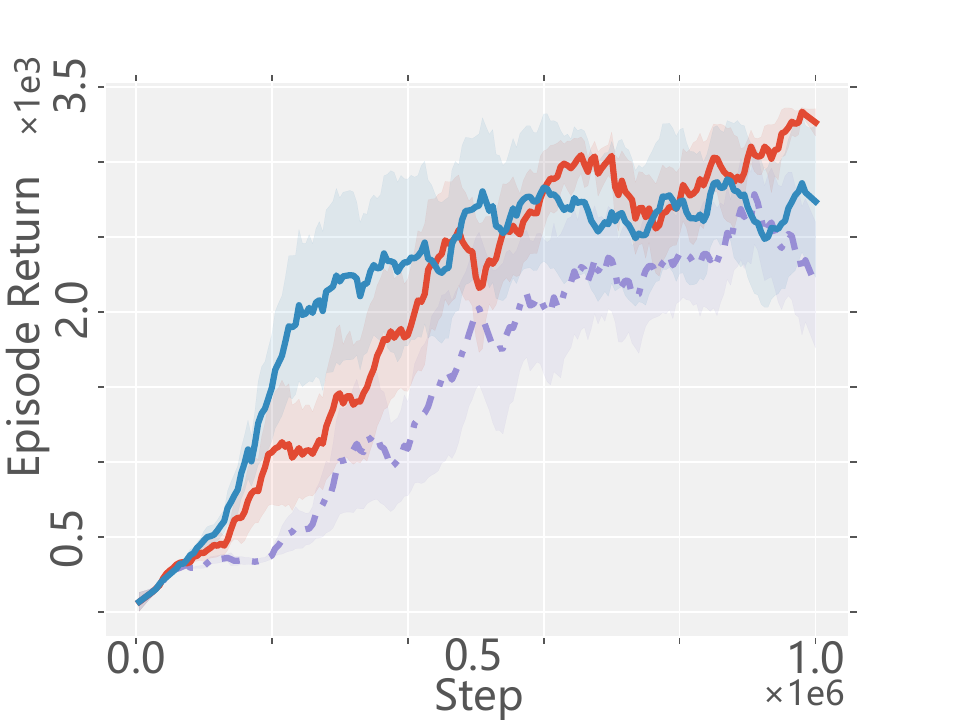}
    \label{Fig:td3-Hopper}
    }
    \hspace{-6.mm}
    \subfloat[Ant-TD3]{
    \includegraphics[width=.38\columnwidth  ]{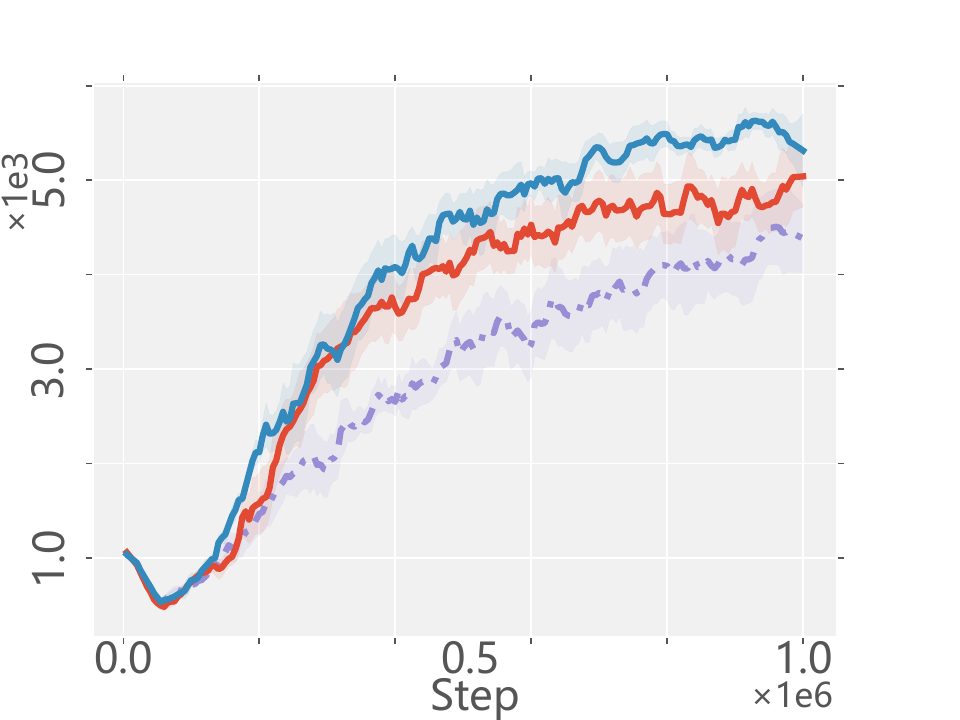}
    \label{Fig:td3-Ant}
    }
    \hspace{-6mm}
    \subfloat[Swimmer-TD3]{
    \includegraphics[width=.38\columnwidth  ]{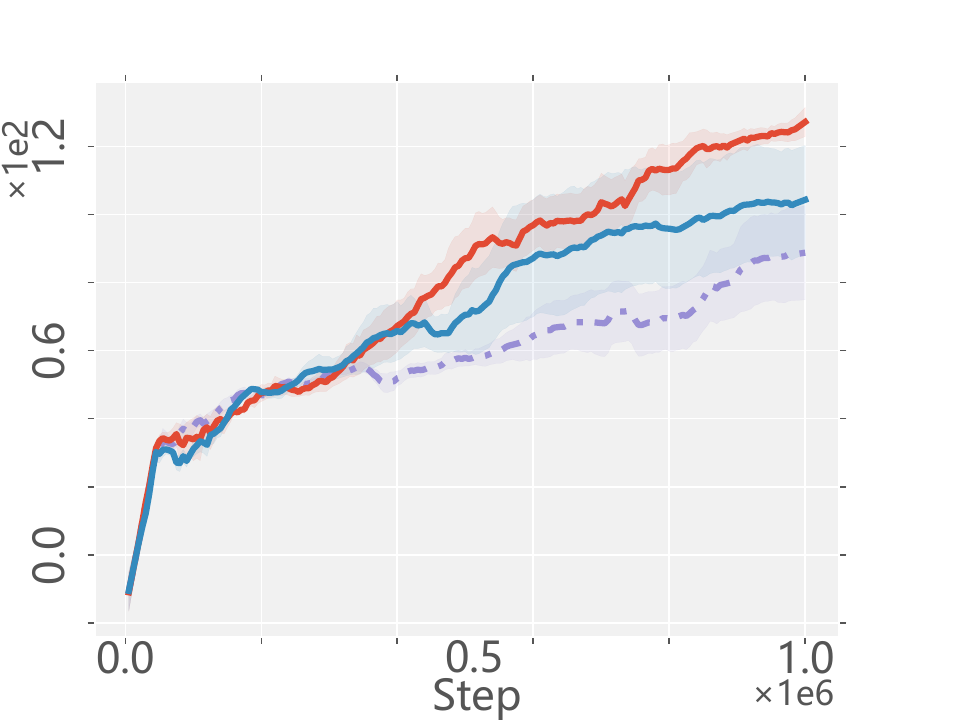}
    \label{Fig:td3-swimmer}
    }
    \hspace{-6mm}
    \subfloat[HalfCheetah-TD3]{
    \includegraphics[width=.38\columnwidth  ]{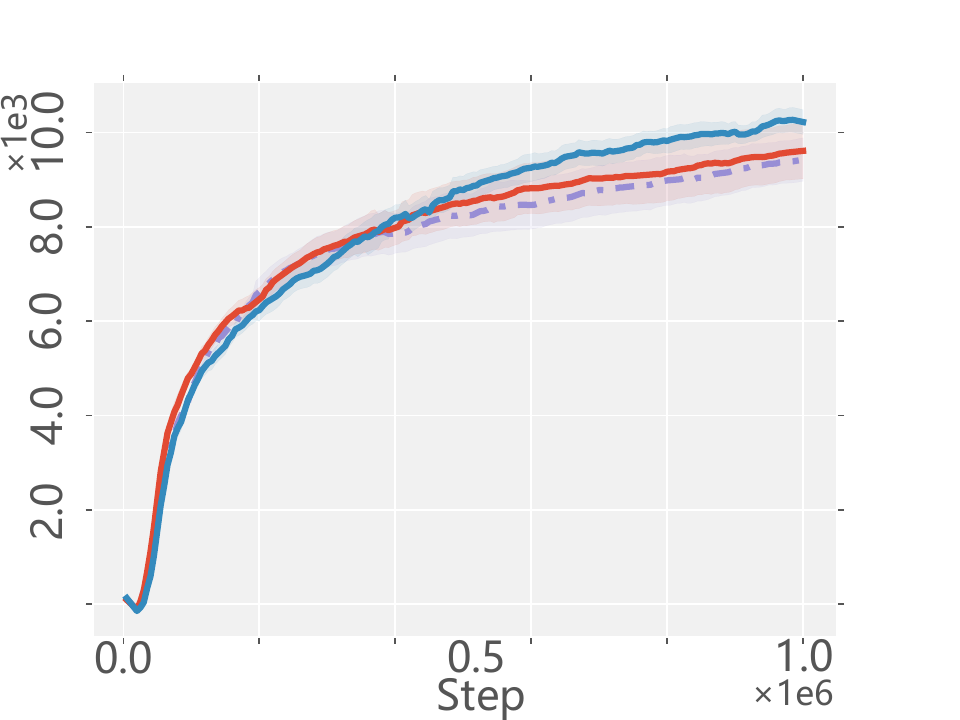}
    \label{Fig:td3-HalfCheetah}
    }
    \hspace{-6mm}
     \subfloat[Humanoid-TD3]{
    \includegraphics[width=.38\columnwidth  ]{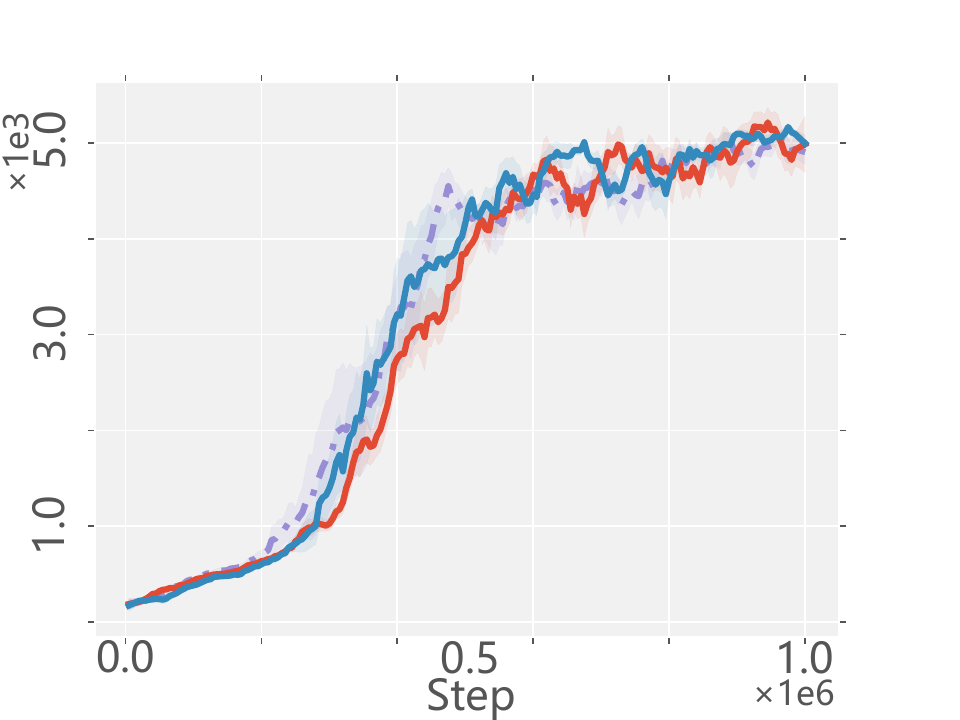}
    \label{Fig:td3-Humanoid}
    }
    \hspace{-6mm}
    \subfloat[Walker2d-TD3]{
    \includegraphics[width=.38\columnwidth  ]{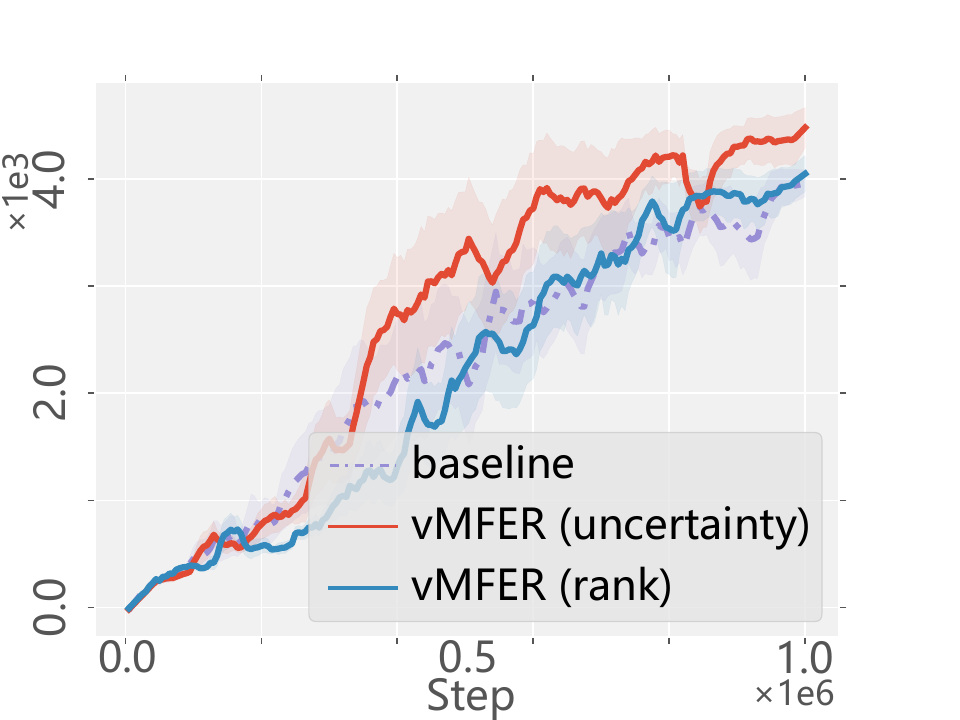}
    \label{Fig:td3-Walker}
    }
    \hspace{-8mm}
    
    \hspace{-6mm} 
    \subfloat[Hopper-SAC]{
    \includegraphics[width=.38\columnwidth  ]{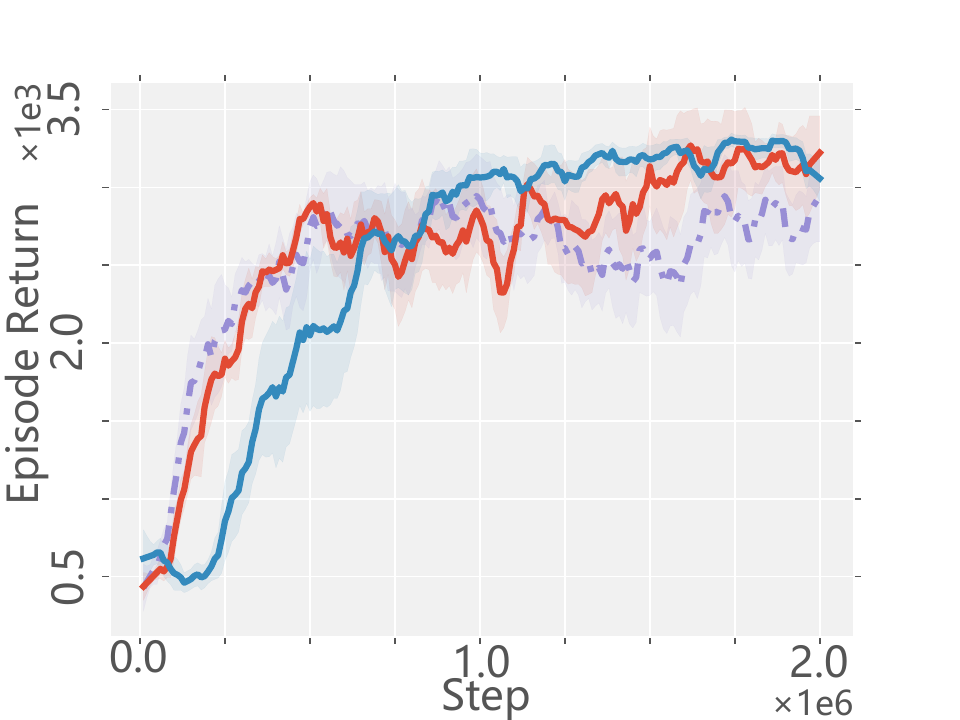}
    \label{Fig:sac-Hopper}
    }
    \hspace{-6mm}
    \subfloat[Ant-SAC]{
    \includegraphics[width=.38\columnwidth ]{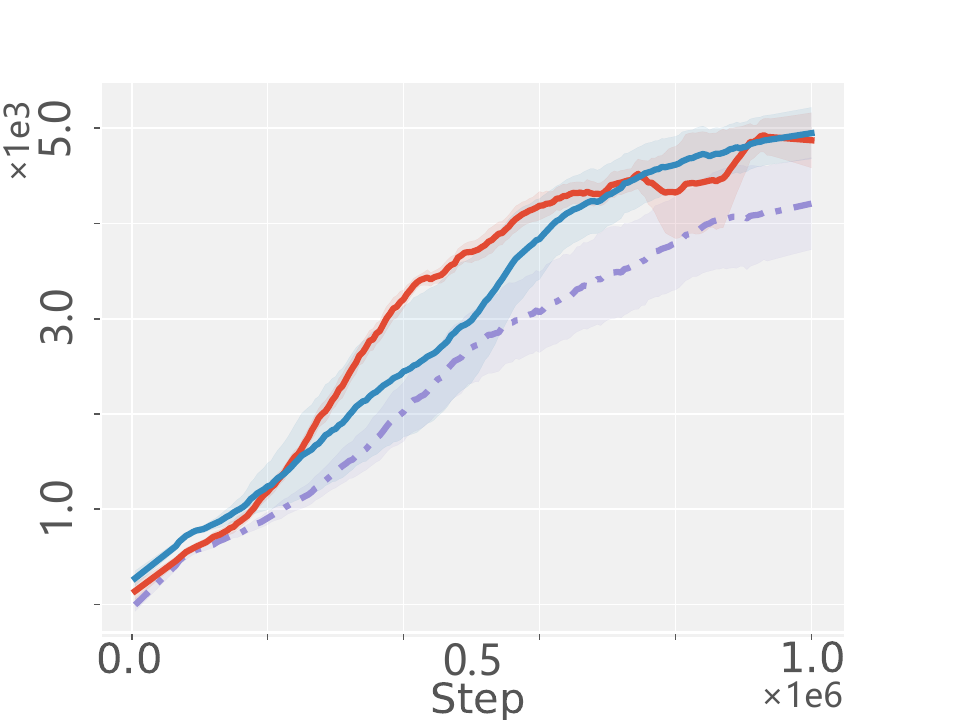}
    \label{Fig:sac-Ant}
    }
   \hspace{-6mm}
    \subfloat[Swimmer-SAC]{
    \includegraphics[width=.38\columnwidth  ]{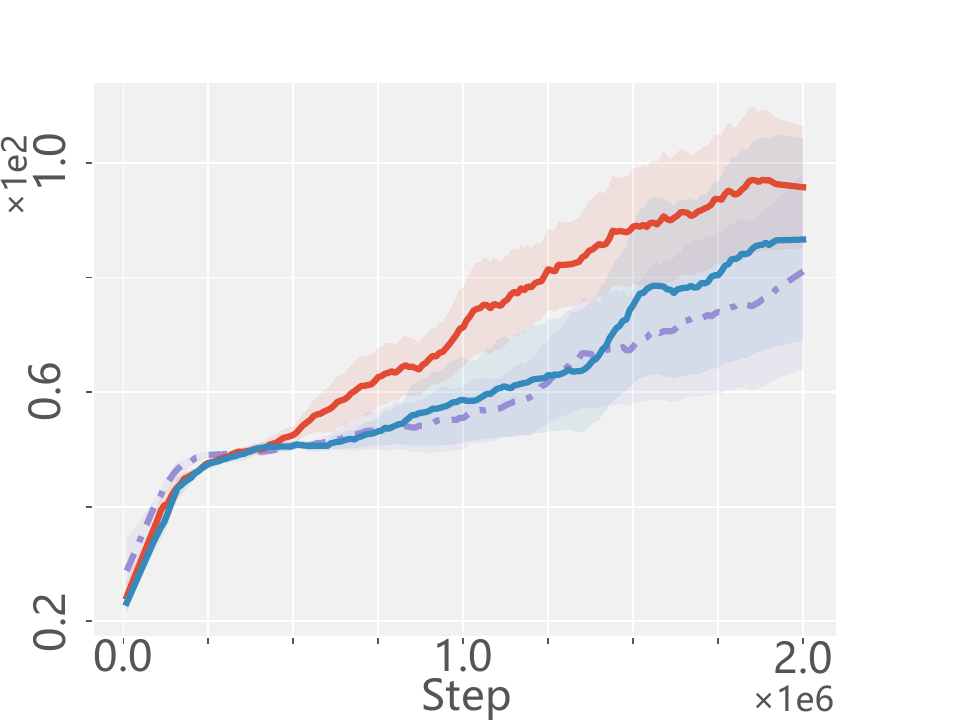}
    \label{Fig:sac-swimmer}
    }
    \hspace{-6mm}
    \subfloat[HalfCheetah-SAC]{
    \includegraphics[width=.38\columnwidth]{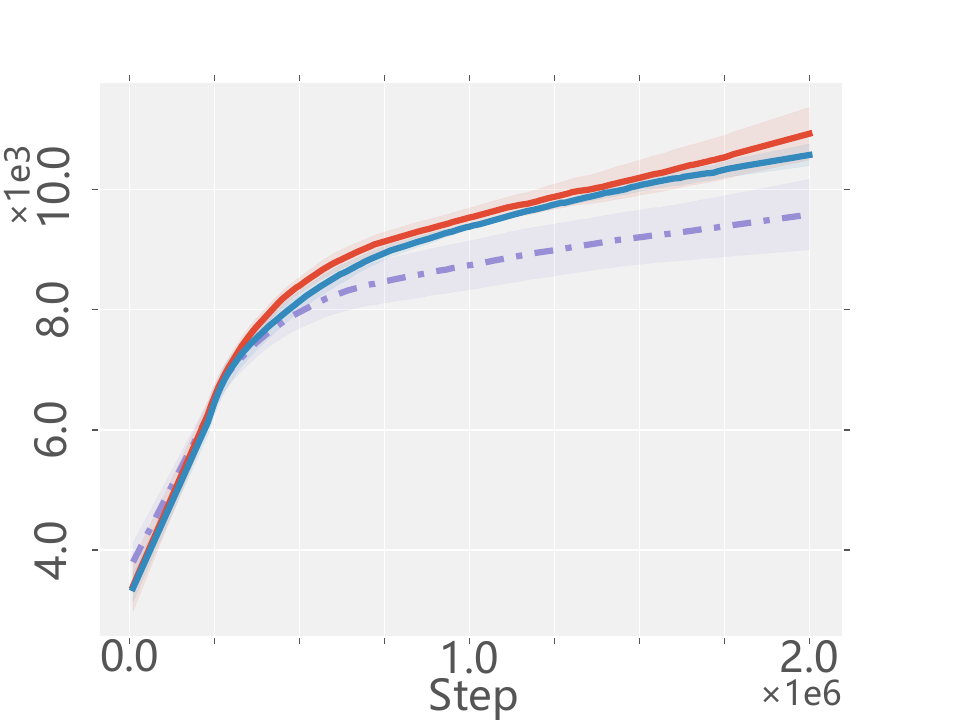}
    \label{Fig:sac-HalfCheetah}
    }
    \hspace{-6mm}
     \subfloat[Humanoid-SAC]{
    \includegraphics[width=.38\columnwidth]{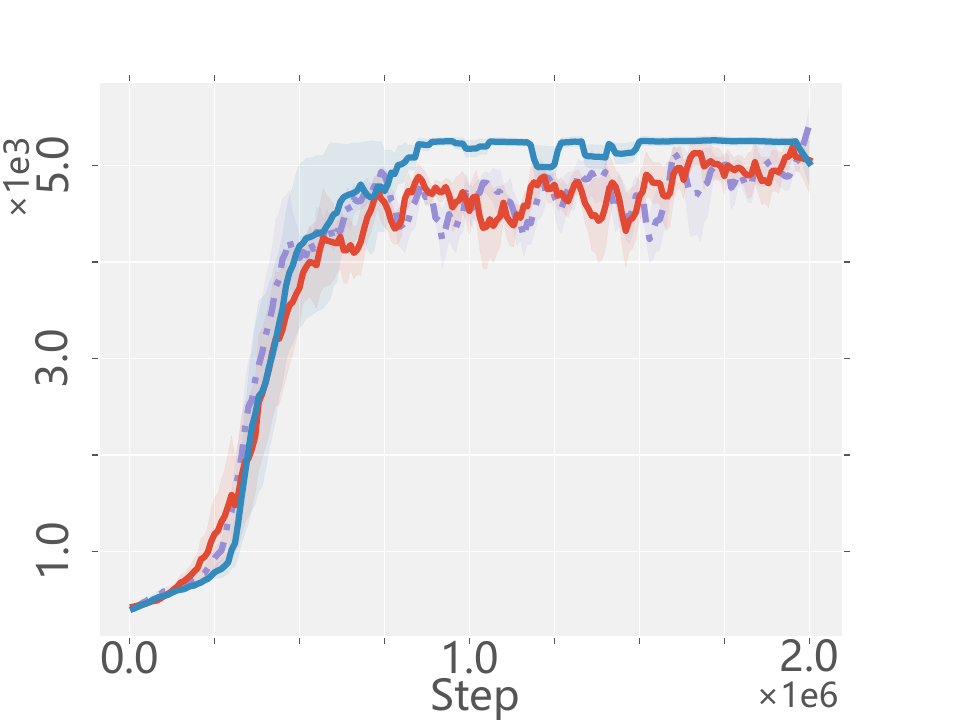}
    \label{Fig:sac-Humanoid}
    }
    \hspace{-6mm}
    \subfloat[Walker2d-SAC]{
    \includegraphics[width=.38\columnwidth]{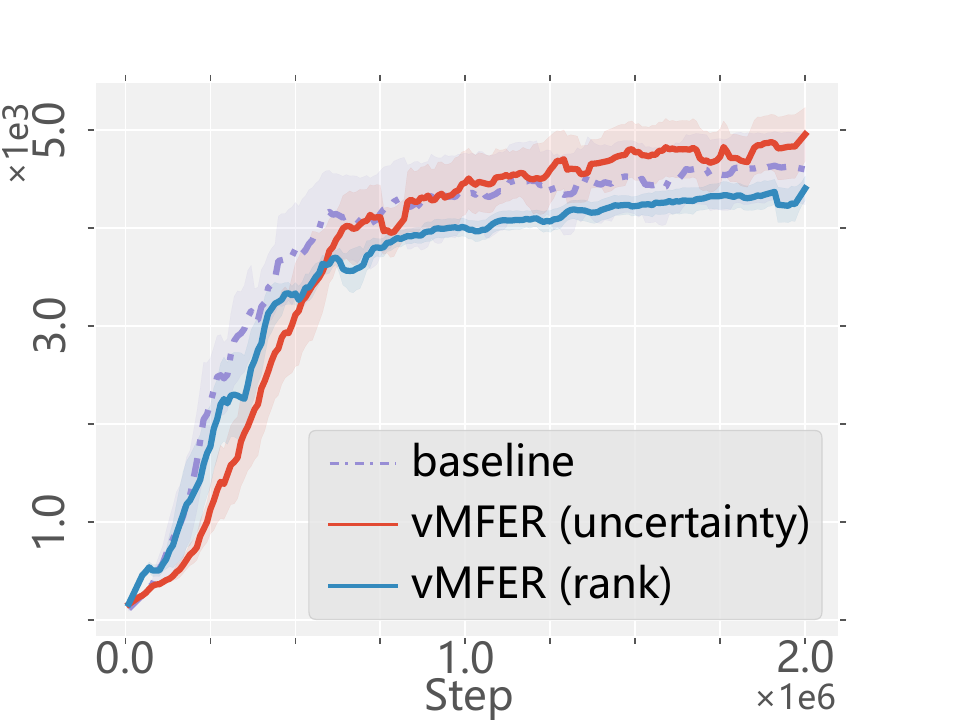}
    \label{Fig:sac-Walker}
    }
    \hspace{-7mm}
    \caption{Examining the performance of vMFER on the Mujoco environment. The baseline curves represents pure TD3 or SAC, while vMFER (uncertainty) and vMFER (rank) represent two distinct forms of vMFER utilized in policy improvement combined with baseline.}
\label{Fig:mujoco+td3}
\end{figure*}

The second approach involves an indirect rank-based method. Here, the probability of the sampling transition is calculated as shown in Eq. ({\ref{Eq:rankprob}}). The rank of transition $(s_j,a_j,r_j,s'_j)$ is determined by sorting the replay memory based on $\exp(\mathbf{R}(s_j)\mu^{\text{T}}(s_j) x(s_j))$ from high to low.
\begin{align}
\small
\begin{split}
     P(j|x(s_j),\mathcal{D}) = \frac{{rank{(\exp(\mathbf{R}(s_j)\mu^{\text{T}}(s_j) x(s_j)))}^{-1}}}{\sum\limits_i^M {rank(\exp(\mathbf{R}(s_i)\mu^{\text{T}}(s_i) x(s_i)))^{-1}}}.
\end{split}
\label{Eq:rankprob}
\end{align}
Our method seamlessly integrates with any Actor-Critic framework algorithm, given that the critic network employs an ensemble structure. Taking TD3 as an example, we combine it with our approach, as detailed in Algorithm \ref{Algo:vMFER_td3}. The key modifications in our enhanced version, marked in brown in Algorithm \ref{Algo:vMFER_td3}, primarily optimize the policy improvement process. The probability of resampling each transition is guided by the uncertainty of gradient directions, and these resampled transitions are subsequently used for policy improvement, aiding in determining the actor's update direction.

It is evident that our method can act as a versatile plugin, which, through Algorithm \ref{Algo:usfpj}, updates and maintains the sampling factor $p_j$ for each transition. This allows for effective resampling during the policy improvement process.

\section{Experimental Results}\label{sec:Exp}

We conducted a series of experiments to evaluate the effectiveness of our vMFER algorithm when combined with off-policy algorithms like TD3 and SAC. 
We aim to compare the performance between different RL methods that incorporate uncertainty-based and rank-based probability updating strategies in a variety of environments. 
Furthermore, we have integrated vMFER with Prioritized Experience Replay (PER) \cite{schaul2015prioritized}, based on SAC, to showcase the flexibility and compatibility of our method. These experiments are mainly centered around the Mujoco robotic control environment \cite{brockman2016openai}.

Additionally, to enhance learning efficiency in sparse reward scenarios, we have merged vMFER with Hindsight Experience Replay \cite{andrychowicz2017hindsight}, achieving notable results in robotic arm control tasks with sparse rewards \cite{plappert2018multi}.
We also conducted ablation studies on the impact of the update-to-data (UTD) ratio \cite{chen2021randomized} on our algorithm, and found that utilizing vMFER could further improve the performance of the algorithm with different UTD ratio values, demonstrating the compatibility between vMFER and UTD ratio.

\paragraph{Implementation Details.}

Apart from hyperparameters associated with baseline algorithms, like ensemble numbers, vMFER requires no fine-tuning of hyperparameters. This facilitates its seamless and efficient integration with any Actor-Critic algorithm to enhance performance.  
We follow the hyperparameter configurations specified in the respective papers of the baseline algorithms.
Besides, the reported results are based on 5 trials, with curves representing means and shaded areas denoting variances.

\paragraph{Performance Improvement.}
\begin{figure*}[!htbp]
\centering
    \hspace{-6mm}
    \subfloat[Hopper]{
    \includegraphics[width=.38\columnwidth  ]{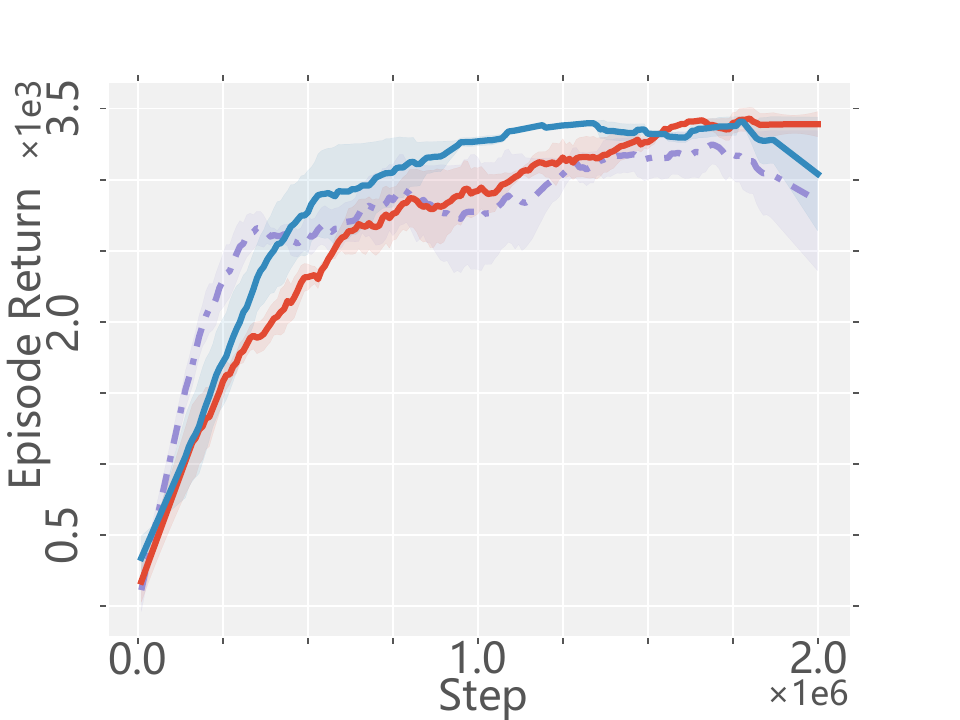}
    \label{Fig:persac-Hopper}
    }
    \hspace{-6mm}
    \subfloat[Ant]{
    \includegraphics[width=.38\columnwidth ]{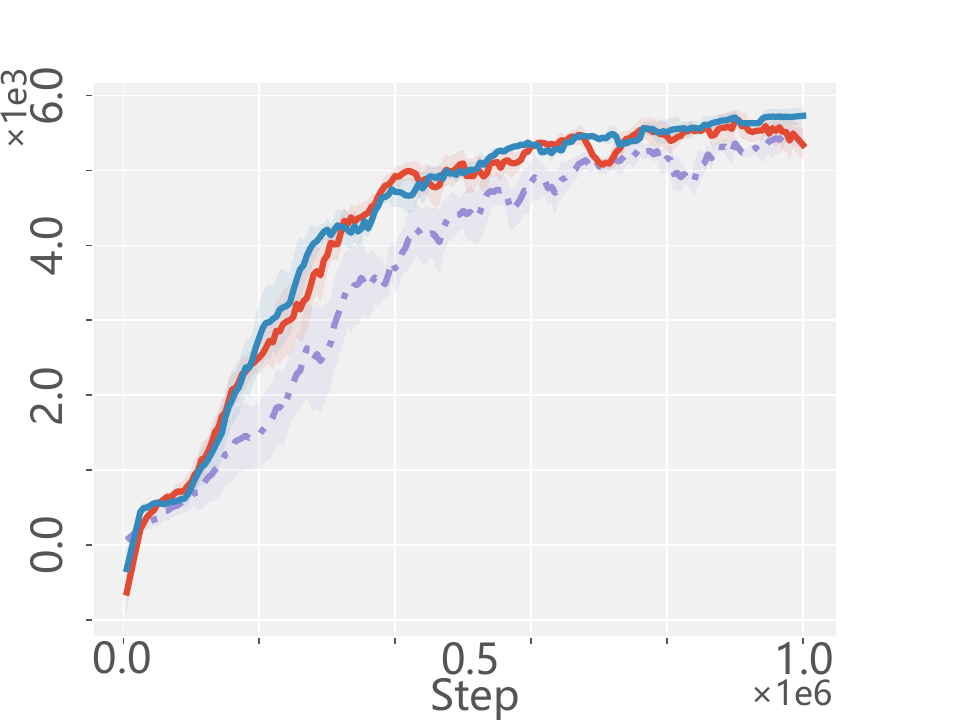}
    \label{Fig:persac-Ant}
    }
     \hspace{-6mm}
    \subfloat[Swimmer]{
    \includegraphics[width=.38\columnwidth  ]{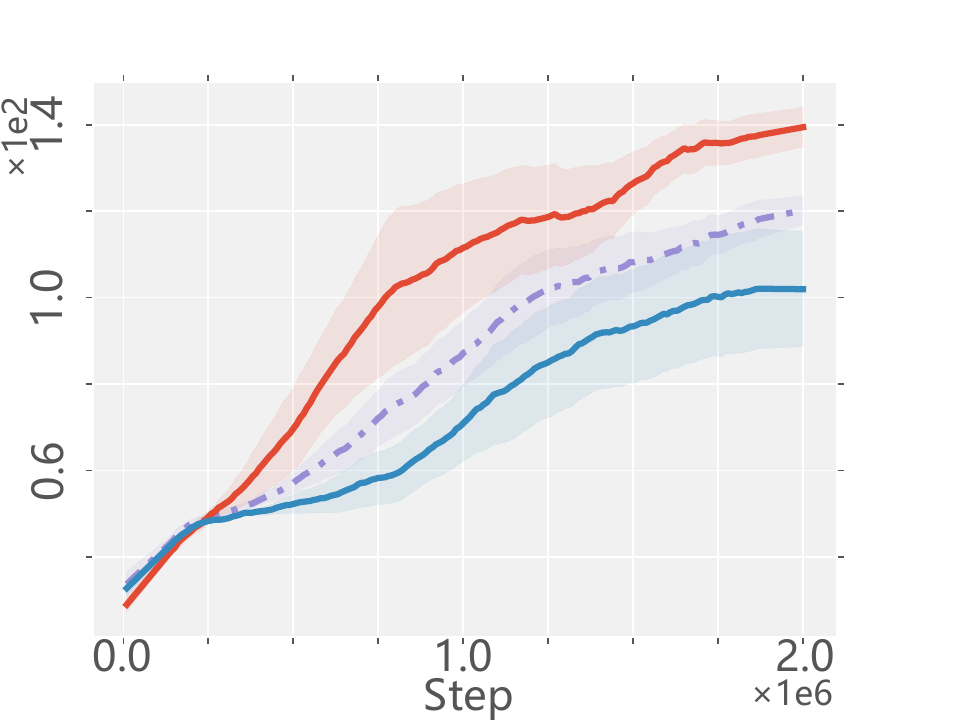}
    \label{Fig:persac-swimmer}
    }
    \hspace{-6mm}
    \subfloat[HalfCheetah]{
    \includegraphics[width=.38\columnwidth]{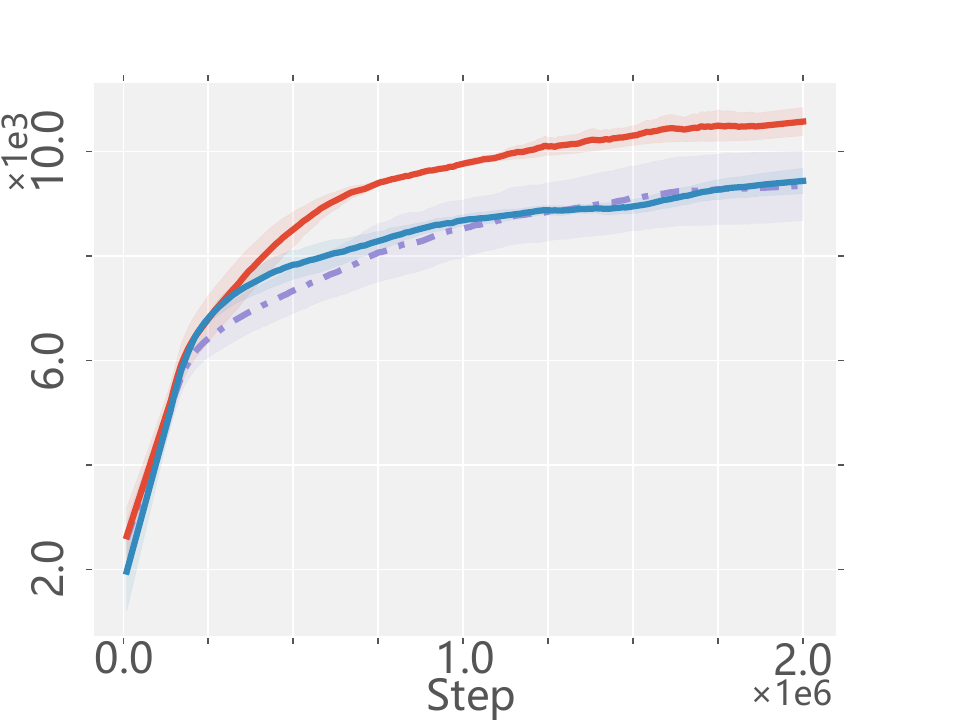}
    \label{Fig:persac-HalfCheetah}
    }
    \hspace{-6mm}
     \subfloat[Humanoid]{
    \includegraphics[width=.38\columnwidth]{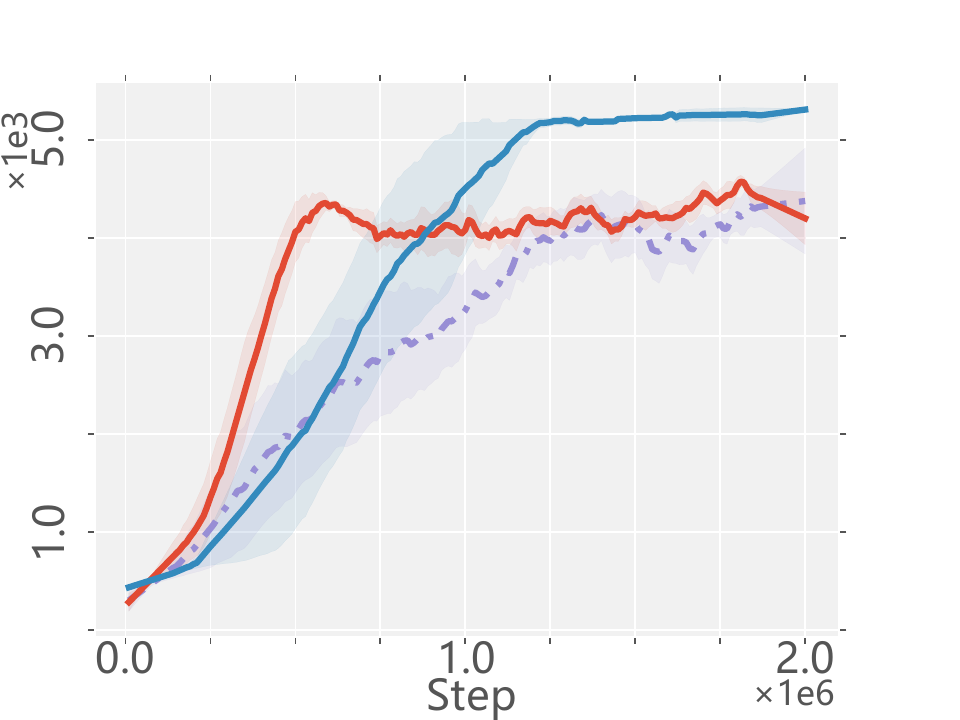}
    \label{Fig:persac-humanoid}
    }
    \hspace{-6mm}
    \subfloat[Walker2d]{
    \includegraphics[width=.38\columnwidth]{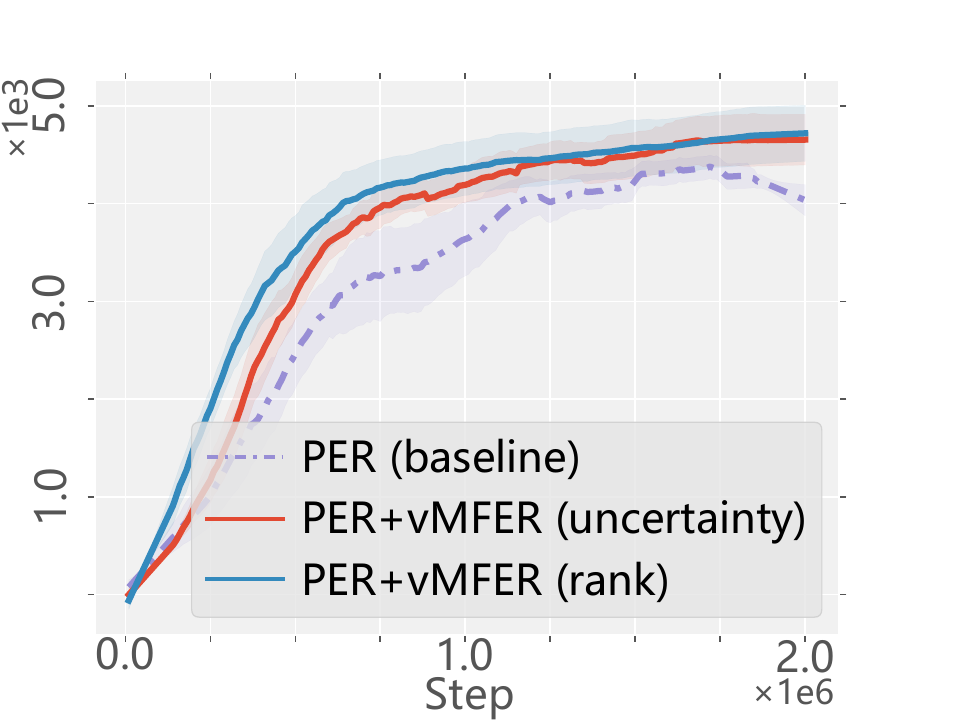}
    \label{Fig:persac-Walker}
    }
    \hspace{-7mm}
\caption{ An experiment conducted in the Mujoco environment to explore the effect of various forms of VMFER on policy improvement. The experiment builds upon the PER combined with SAC.}
\label{Fig:mujoco+persac}
\end{figure*}

Results presented in Figure \ref{Fig:mujoco+td3} demonstrate significant performance improvement for both TD3 and SAC algorithms compared to their respective baseline algorithms.
Furthermore, in Figure \ref{Fig:mujoco+persac}, we investigate 
the influence of vMFER on the combination of PER, based on SAC. 
Table \ref{tab:my_label} presents the average performance improvement achieved by integrating vMFER with various algorithms (SAC, TD3, PER) in Mujoco tasks, compared to their baseline counterparts.

\begin{table}[t]
    \centering
    \small
    \begin{tabular}{l|ccc}
    \toprule
     &  SAC & TD3 & SAC+PER \\
    \midrule
        baseline & 100\%& 100\%& 100\%\\
         vMFER {\small(rank)} & 106.84\% & 111.62\% & 102.09\%\\
         vMFER {\small(uncertainty )} & \textbf{113.78}\% & \textbf{117.75}\% & \textbf{107.17}\%\\
    \bottomrule
    \end{tabular}
    \caption{
    Average performance improvement of vMFER over baseline, calculated by aggregating performance gains across all tasks.}
    \label{tab:my_label}
\end{table}
Irrespective of whether the resampling probability of transition was updated directly through uncertainty or rank, the overall performance is notably superior to the baseline.
In various environments, vMFER exhibits distinct performance enhancements with rank and uncertainty. 
In summary, an average performance improvement of over 10\% compared to the baseline was accomplished.
These findings highlight the importance of avoiding the blind use of transitions during the policy improvement process, which may reduce efficiency.
Our method of reassigning the confidence of transitions by the uncertainty of gradient directions during the policy improvement process is more efficient.

\paragraph{Extended Analyses.}
Our integration of vMFER with PER, known for its transition sampling probability redistribution based on TD error, still yields significant performance improvements. This suggests that both PER and vMFER methods independently exert influence on the SAC algorithm. The superior performance of the combined PER and vMFER algorithm compared to PER alone implies that their effects on the algorithm are somewhat orthogonal. Indeed, while PER primarily optimizes the policy evaluation process, vMFER enhances the policy improvement process.

Moreover, we also find that policy improvement using vMFER, which relies on the uncertainty in transition sampling probability, is more stable and effective than using rank-based methods, challenging the robustness associated with rank in PER. This divergence can be attributed to the finite range of uncertainty in our approach. The vMFER algorithm calculates the sampling factor $p_j$ as {\footnotesize $p_j = \exp(\mathbf{R}\mu^{\text{T}}x)= \exp(\mathbf{R}\cos \xi)$}, where $\xi$ is the angle between the selected gradient of the RL algorithm and the mean direction $\mu$ of the distribution. Here, the uncertainty is quantified on a smaller scale by substituting $k$ with $\mathbf{R}$.
\begin{figure}[t]
\centering
\vspace{-6mm}
\hspace{-6mm}
 \subfloat[Rank]{
    \includegraphics[width=0.54\columnwidth  ]{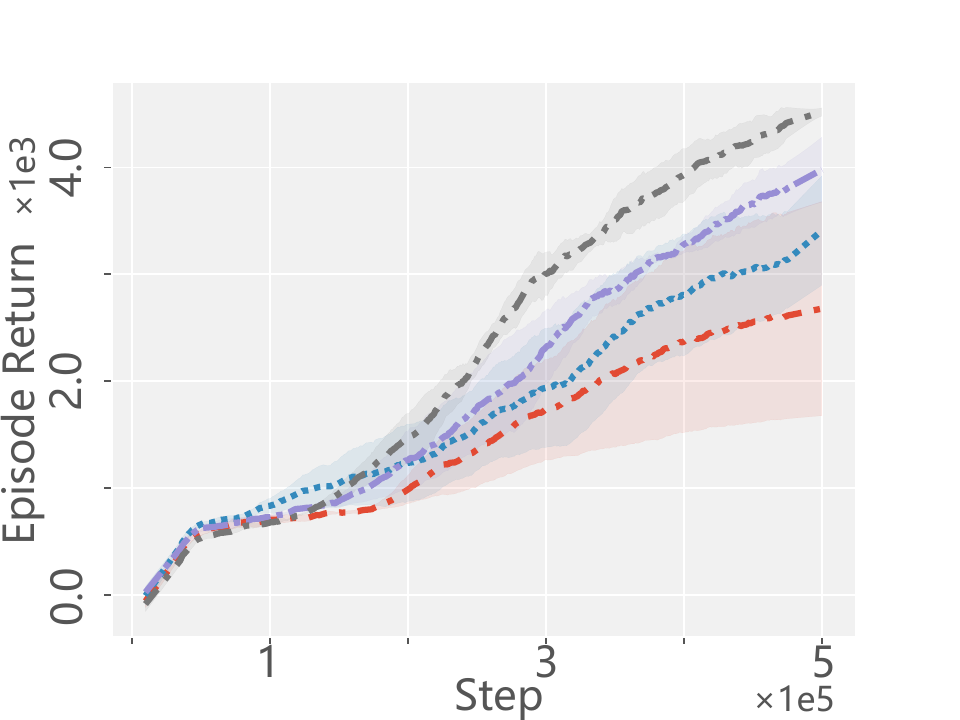}
    \label{Fig:ablationen_rank}
    }
    \hspace{-7mm}
\subfloat[Uncertainty]{
\includegraphics[width=0.54\columnwidth  ]{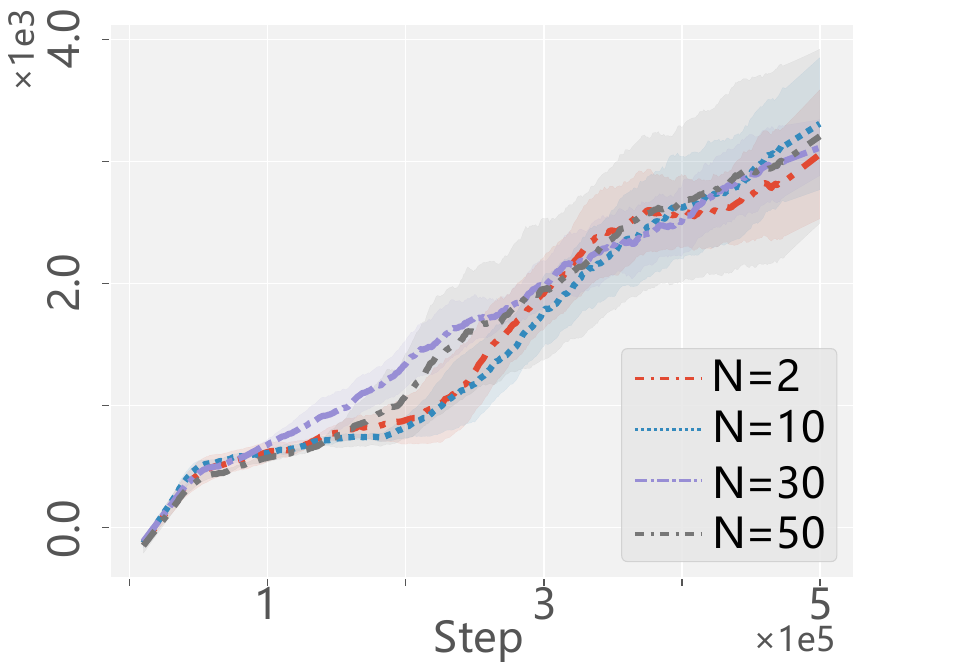}
\label{Fig:ablationen_un}
}
\hspace{-11mm}
\caption{ The impact of ensemble number on vMFER.}
 \label{Fig:ablationen_ens}
\end{figure}
\paragraph{Ablation on Ensemble Number.} In addition,  we also investigate how the use of different numbers of ensemble critics for calculating gradient uncertainty in vMFER influences its performance, as depicted in Figure \ref{Fig:ablationen_ens}.
It is important to emphasize that these additional critics are solely employed to enhance the calculation of gradient uncertainty and do not affect policy evaluation.
Our findings reveal that modifications in ensemble size significantly affect the performance of vMFER when the resampling probability is determined by rank, as illustrated in Figure \ref{Fig:ablationen_ens}\subref{Fig:ablationen_rank}. 
Higher ensemble size results in improved performance. 
However, varying the ensemble size has minimal effect on vMFER when the resampling probability is determined by uncertainty, observed in Figure \ref{Fig:ablationen_ens} \subref{Fig:ablationen_un}.

\section{Related Work}\label{SE:relatedwork}

\paragraph{Ensemble Structure in RL.}
Ensemble structures enhance RL algorithm performance \cite{buckman2018sample,lee2021sunrise,shen2021robust,song2023ensemble,lee2022offline}, addressing overestimation in stochastic MDPs, as seen in Q-learning \cite{watkins1992q}. Double Q-learning \cite{hasselt2010double} initially used dual critics to counteract overestimation.
Averaged Q-estimates by Anschel et al. \cite{anschel2017averaged} reduced Q-learning variance, while Lan et al. \cite{lan2020maxmin} and Ciosek et al. \cite{ciosek2019better} utilized ensembles for exploration and conservative updates. In offline RL, ensemble critics are used for training more stable, or even conservative critics~\cite{agarwal2020optimistic,an2021uncertainty,off2onensemble}.

\paragraph{Uncertainty Measure in RL.} Uncertainty estimation is widely used in RL \cite{chen2017ucb,lockwood2022review,kalweit2017uncertainty,zhang2020robust,clements2019estimating} for exploration \cite{audibert2009exploration,yang2021exploration,OVDExplorer}, Q-learning \cite{dearden1998bayesian,wang2021online}, and planning \cite{wu2022plan}. Bootstrapped DQN \cite{osband2016deep} uses an ensemble of Q-functions for uncertainty quantification in Q-values, enhancing exploration. Osband et al. \cite{osband2018randomized} propose a Q-ensemble with Bayesian prior functions. Abbas et al. \cite{abbas2020selective} introduce uncertainty-incorporated planning with imperfect models. In offline RL, MOPO \cite{yu2020mopo} and MOReL \cite{kidambi2020morel} employ model prediction uncertainty measures to address uncertainty-penalized policy optimization.

\section{Conclusion}
\label{sec:conclusion}
We have advanced the policy improvement process by incorporating the consideration of gradient direction disagreements under an ensemble structure. Distinct from prior methodologies, our approach utilizes von Mises-Fisher distributions to model gradient directions and quantify the uncertainty of these directions under current critics for each transition during policy improvement. Building on this, we introduce the vMFER algorithm, which assigns confidence levels to all transitions in the replay buffer and resamples them based on their probability, determined by the uncertainty of gradient directions. In this way, the transition with high confidence can be used to update actors more frequently, thereby enhancing the efficiency of the policy improvement process.

The impact of gradient uncertainty on the policy improvement process, considered in this paper, is an aspect that has been scarcely addressed in existing research. This insight prompts future researchers to be aware of the potential effects of ensemble gradients. Future studies could delve deeper into the uncertainty of gradients, extending from solely directional uncertainty to the joint uncertainty of both direction and magnitude. Furthermore, exploring the quantification of gradient uncertainty, its impact in offline RL, and its advantages in practical implementations holds substantial value.

\clearpage

\bibliographystyle{named}
\bibliography{ijcai24}

\appendix
\onecolumn
\section{Additional Clarifications on Eq. (\ref{Eq:prob})}\label{Appdix:kandR}%Monotonicity of Concentration Parameter $k$
In Eq. (\ref{Eq:Baye}), we introduce the parameter $k$ as the concentration parameter of the von Mises-Fisher distribution, with $\hat{k}$ serving as an approximation of $k$.
However, the action's dimensionality greatly influences the value range of $\hat{k}$, resulting in significant variations between its maximum and minimum values.
Using such a parameter to measure the uncertainty of gradient directions might lead to an overdependence on a limited number of transitions during resampling, failing to fully utilize all transitions in the replay buffer. 
Consequently, we propose employing $\mathbf{R}$ as an alternative to $\hat{k}$ for quantifying the uncertainty of gradient directions. To demonstrate the feasibility of replacing $\hat{k}$ with $\mathbf{R}$, this section analyzes the monotonic relationship between $\hat{k}$ and $\mathbf{R}$.

By calculating $\frac{\partial \hat{k}}{\partial\mathbf{R}}$, we find:
\begin{align}
\begin{split}
       \frac{\partial \hat{k}}{\partial\mathbf{R}}& = \frac{\mathbf{R}^4+(p-3)\mathbf{R}^2 +p}{(1-\mathbf{R}^2)^2} \qquad \mathbf{R} \in [0,1]\\
       &\left\{ 
    \begin{array}{lc}
         \text{if}\quad  p=1 \quad\frac{\partial \hat{k}}{\partial\mathbf{R}} =|1-R^2|>0\\
          \text{if}\quad  p=2 \quad\frac{\partial \hat{k}}{\partial\mathbf{R}} = \frac{(\mathbf{R}^2-\frac{1}{2})^2+\frac{7}{4}}{(1-\mathbf{R}^2)^2}>0\\
           \text{if}\quad  p\geq 3 \quad \frac{\partial \hat{k}}{\partial\mathbf{R}}\geq \frac{\mathbf{R}^4+p}{(1-\mathbf{R}^2)^2}>0
    \end{array}
\right.
\end{split}
\end{align}
This reveals that $\mathbf{R}$ is directly proportional to $\hat{k}$, suggesting that an increase in $\hat{k}$ aligns with an increase in $\mathbf{R}$, thereby validating the use of $\mathbf{R}$ as an alternative to $\hat{k}$.

\section{The Expression of $C_p(k)$ } \label{app:c_pk}
In Eq. (\ref{Eq：vmf_pdf}), we introduce $C_p(k)$ to denote the normalization constant, given by \cite{sra2012short}:
\begin{align}
    C_p(k)=\frac{k^{p/2-1}}{(2\pi)^{p/2} I_{p/2-1}(k)}
\end{align}
Here, $I_{p/2-1}$ represents the modified Bessel function of the first kind at order $p/2-1$, which is expressed as:
\begin{align}
    I_{p/2-1}(k) = \sum\limits_{m \geq 0} \frac{1}{\Gamma(p/2+m) m!}(\frac{k}{2})^{2m+p/2-1}
\end{align}
where $\Gamma(\cdot)$ refers to the well-known Gamma function.

\section{The Definition of sampling factor $p_j$}

In Section \ref{SE:Howtousevmf}, we introduced a corresponding sampling factor $p_j$ for each transition $(s_j,a_j,r_j,s'_j)$, and utilized {\footnotesize $P(j| x(s_j),\mathcal{D}) = \frac{p_j}{\sum_m p_m}$} to represent the posterior probability distribution of sampling a particular transition.
 To comprehensively demonstrate how Eq. (\ref{Eq:Baye}) and Eq. (\ref{Eq:prob}) were derived, this section will sequentially delve into the derivations of Prior Probability, Conditional Probability, and Posterior Probability.
Additionally, we also explore substituting $p_j$ with $\hat{p}_j$ to reduce computational complexity and provide a proof of the asymptotic equivalence between $\hat{p}_j$ and $p_j$.
Finally, to mitigate the shortcomings associated with the concentration parameter $k$, as described in Appendix \ref{Appdix:kandR}, we ultimately opt for $\Bar{p}_j$ as a substitute for $\hat{p}_j$.

\paragraph{Prior Probability.}

We denote $b$ as the number of transitions sampled from the replay buffer. The prior probability of sampling these $b$ transitions can be calculated as:
\begin{align}
    \begin{split}
        P(\{s_j\}_{j=1}^b|\mathcal{D}) &= P(s_1|\{s_j\}_{j=2}^b,\mathcal{D})
        P(s_2|\{s_j\}_{j=3}^b,\mathcal{D})\cdot\cdot\cdot
        P(s_M|\mathcal{D})\\
        &= \prod_{j=1}^b{P(s_j|\mathcal{D})} =  \prod_{j=1}^b{P(j|\mathcal{D})}\\
    \end{split}
\end{align}
\paragraph{Conditional Probability.}

Next, we use $g_j = \frac{\partial l(s_j,a)}{\partial a}\frac{\partial a}{ \partial \theta}|_{a\sim\pi_\theta(\cdot|s_j)}  $, where $\theta$ represents the parameters of the actor network, to describe the gradient contributed by transition $s_j$ to the actor's update, as detailed in Eq. (\ref{Eq:losses}). Therefore, the conditional probability distribution can be expressed as:
\begin{align}
    \begin{split}
        P(\{g_j\}_{j=1}^b|\{s_j\}_{j=1}^b,\mathcal{D}) & = P(g_1|\{s_j\}_{j=1}^b,\mathcal{D}) P(\{g_j\}_{j=2}^b|g_1,\{s_j\}_{j=1}^b,\mathcal{D})\\
    &=\prod_{j=1}^{b} P(g_{j}|\{g_m\}_{m=1}^{j-1},\{s_m\}_{m=1}^{b},\mathcal{D})
    =\prod_{j=1}^{b} P(g_{j}|\{s_m\}_{m=1}^{b},\mathcal{D})
    \end{split}
    \label{Eq:_p(g|s)}
\end{align}

Given the high dimensionality of $g_j$, which often leads to sparsity in high-dimensional vectors, we propose the following assumption:
\begin{Assumption}
$g_j$ is solely related to $s_j$ and does not consider correlations with $\{s_m\}_{m\neq j}$, that is,
\begin{align*}
     P(g_{j}|\{s_m\}_{m=1}^{b},\mathcal{D}) =P(g_{j}|s_j,\mathcal{D})
\end{align*}
\end{Assumption}
Therefore, Eq. (\ref{Eq:_p(g|s)}) can be further simplified to:
\begin{align}
    \begin{split}
        P(\{g_j\}_{j=1}^b|\{s_j\}_{j=1}^b,\mathcal{D}) 
    =\prod_{j=1}^{b} P(g_{j}|s_j,\mathcal{D}) = \prod_{j=1}^{b} P( \frac{\partial l(s_j,a)}{\partial a}\frac{\partial a}{ \partial\theta}|s_j,\mathcal{D}), \;{a\sim\pi_\theta(\cdot|s_j)}
    \end{split}
    \label{Eq:p(g|s)}
\end{align}

Furthermore, as indicated in Eq. (\ref{Eq:chainrule}), during actor's update, $\pi_{\theta}$ consistently generates the same action $a$ for a given transition $s_j$. Consequently, $\frac{\partial a}{\partial \theta}|_{a\sim \pi(\cdot|s_j)}$ can be regarded as a constant. By considering the gradient induced by $s_j$ as adhering to a specific distribution, we can infer the following:
\begin{align}
    \begin{split}
      P( \frac{\partial l(s_j,a)}{\partial a}\frac{\partial a}{ \partial \theta}|s_j,\mathcal{D}) \propto P( \frac{\partial l(s_j,a)}{\partial a}|s_j,\mathcal{D}), \;{a\sim\pi_\theta(\cdot|s_j)}
    \end{split}
    \label{Eq:p(g|s)}
\end{align}

Our study primarily focuses on the uncertainty of gradient directions, as demonstrated in Section \ref{SE:howtomeasure}, we 
use {\footnotesize$x(s_j) = ||\frac{\partial l(s_j,a)}{\partial a}||_2^{-1} \boldsymbol{\cdot} \frac{\partial l(s_j,a)}{\partial a} |_{a\sim\pi(\cdot|s_t)}\; $} to denote the direction of the gradient contributed by transition $s_j$ to the actor's update.
And we consider that $x(s_j)$ is a gradient direction that sampled from $\text{vMF}(k(s_j),\mu(s_j))$, described in Section \ref{SE:Howtousevmf}.
Similarly, we can derive the conditional probability distribution of $x(s_j)$ as follows:
\begin{align}
    \begin{split}
        P(\{x(s_j)\}_{j=1}^b|\{s_j\}_{j=1}^b,\mathcal{D}) 
    =\prod_{j=1}^{b} P(x(s_j)|s_j,\mathcal{D}) 
    \end{split}
    \label{Eq:p(g|s)}
\end{align}

\paragraph{Posterior Probability.}
As described in Section $\ref{SE:Howtousevmf}$, we calculate the posterior probability distribution of different transitions $s_j$ being sampled based on the gradient directions they contribute to the actor's update, as follow:
\begin{align}
    \begin{split}
        &P(\{s_j\}_{j=1}^b|\mathcal{D}) P(\{x(s_j)\}_{j=1}^b|\{s_i\}_{i=1}^b,\mathcal{D}) \propto  P(\{s_j\}_{j=1}^b|\{x(s_j)\}_{j=1}^b,\mathcal{D})\\
         \xrightarrow{} & \prod_{j=1}^b P(s_j|\mathcal{D}) P(x(s_j)|s_j,\mathcal{D}) \propto \prod_{j=1}^b {P(s_j|x(s_j),\mathcal{D})}
    \end{split}
    \label{Eq:prodsx}
\end{align}
Given the parametric form of the posterior probability distribution $\frac{p_j}{\sum_m p_m}$ as specified in Eq. (\ref{Eq:prob}), and considering that this study models the gradient directions generated by transition $s_j$ for the actor's update using the von Mises-Fisher distribution, we can further deduce Eq. (\ref{Eq:prodsx}) as follows:
\begin{align}
    \begin{split}
        \prod \limits_{j=1}^b  \frac{1}{M} C_p(k(s_j))\exp({k(s_j)\mu(s_j)^\text{T}x(s_j)}) \propto
        \prod\limits_{j=1}^{b}\frac{p_j}{\sum\limits_{m=1}^M  p_m}
    \end{split}\label{Eq:priorandposterior}
\end{align}
where $M$ denotes the total number of data in replay buffer $\mathcal{D}$, defined in Section \ref{SE:Howtousevmf}.  
Then, we can establish a posterior probability that satisfies our desired requirements, as mentioned in Eq. ({\ref{Eq:priorandposterior}}), by using the formula shown in Eq. ({\ref{Eq:Baye_def}}).
\begin{align}
% \centering
\begin{split}
p_j &= C_p(k(s_j))exp(k(s_j)\mu(s_j)^{\text{T}}x(s_j))\\
P(s_j|\{x(s_m)\}_{m=1}^b,\mathcal{D}) = P(s_j|x(s_j),\mathcal{D}) &=P(j|x(s_j),\mathcal{D}) 
=\frac{ C_p(k(s_j))exp(k(s_j)\mu(s_j)^{\text{T}}x(s_j))}{\sum\limits_{m=1}^M C_p(k(s_m))exp(k(s_m)\mu(s_m)^{\text{T}}x(s_m))}
\label{Eq:Baye_def}
\end{split}
\end{align}
Ideally, we would update the sampling factor $p_j$ for all transitions in the replay buffer after each actor update. However, due to computational resource considerations, we update only $b$ sampling factors each time.

\paragraph{Reducing Computational Complexity.}
Calculating $C_p(k(s_j))$ is computationally expensive. Therefore, we propose a more computationally efficient formalism, as shown in Eq. ({\ref{Eq:Baye_def_hat}}).
\begin{align}\begin{split}
\hat{p}_j &= exp(k(s_j)\mu(s_j)^{\text{T}}x(s_j))\\
\hat{P}(j|x(s_j),\mathcal{D}) &=\frac{ exp(k(s_j)\mu(s_j)^{\text{T}}x(s_j))}{\sum\limits_{m=1}^M exp(k(s_m)\mu(s_m)^{\text{T}}x(s_m))}
\label{Eq:Baye_def_hat}
\end{split}
\end{align}

\paragraph{Asymptotic Equivalence.} We can then obtain the relationship between $\hat{P}(j|x(s_j),\mathcal{D})$ and $P(j|x(s_j),\mathcal{D})$ easily, as demonstrated in Eq. ({\ref{Eq:relationofP}}).
\begin{align}\begin{split}
\frac{\min_j C_p(k(s_j)) }{\max_j C_p(k(s_j))} \hat{P}(j|x(s_j),\mathcal{D})
\leq{P}(j|x(s_j),\mathcal{D})
\leq\frac{\max_j C_p(k(s_j)) }{\min_j C_p(k(s_j))} \hat{P}(j|x(s_j),\mathcal{D})
\label{Eq:relationofP}
\end{split}
\end{align}
Since we know that $C_p(k(s_j))$ is the normalization constant of the von Mises-Fisher distribution for each transition, both $\frac{\max_j C_p(k(s_j)) }{\min_j C_p(k(s_j))}$ and $\frac{\min_j C_p(k(s_j)) }{\max_j C_p(k(s_j))}$ approach 1 as the training progresses. 

Moreover, to avoid potential polarization of $k$ calculation due to the denominator $1-\mathbf{R}^2(s_j)$ approaching 0, and reduce computational cost, we utilize $\mathbf{R}$ instead of $k$ (as proven in Appendix \ref{Appdix:kandR} that $k\propto \mathbf{R}$). The final formalism of the posterior probability is shown in Eq. ({\ref{Eq:finalp}}).
\begin{align}\begin{split}
\Bar{p}_j &= exp(\mathbf{R}(s_j)\mu(s_j)^{\text{T}}x(s_j))\\
\Bar{P}(j|x(s_j),\mathcal{D}) &=\frac{ exp(\mathbf{R}(s_j)\mu(s_j)^{\text{T}}x(s_j))}{\sum\limits_{i=1}^M exp(\mathbf{R}(s_i)\mu(s_i)^{\text{T}}x(s_i))}
\label{Eq:finalp}
\end{split}
\end{align}

In the actual implementation of the algorithm, we compute $\Bar{p}_j$ as an alternative to calculating $p_j$. We then apply the approach outlined in Eq. (\ref{Eq:finalp}) to compute the posterior probability distribution, which aligns with the mathematical formulation illustrated in Eq. (\ref{Eq:prob}).

\section{Exploring the Connection: vMF Distribution and Cosine Similarity.}\label{APP:vmf&cos}

In Sections \ref{SE:Asimpleexample} and \ref{SE:NecOFun}, we utilized the angle between gradients to quantify the uncertainty of gradient directions for a clear presentation of the Simple Example and Toy Experiment effects. However, in Section \ref{SE:howtomeasure}, we adopted the von Mises-Fisher distribution for modeling gradient directions. In this section, we present the correlation between these two methods of measuring uncertainty when the ensemble number is 2, further elucidating the advantages of using the von Mises-Fisher distribution.

Utilizing Banerjee's method allows us to estimate the concentration parameter $k$ and mean direction $\mu$, as delineated in Eq.(\ref{Eq:rmuk}). By setting the ensemble number to 2, as discussed in Section~\ref{SE:Asimpleexample} and Section~\ref{SE:NecOFun}, and representing the angle between gradients with ${\Theta} \in [0^\circ, 180^\circ]$, we derive the following:
\begin{align}
\begin{split}
        \mathbf{x}(s_t) &= \frac{x_1(s_t) + x_2(s_t)}{2}\\
        \mathbf{R}(s_t) &= ||\mathbf{x}(s_t) ||_2 = \sqrt{(\frac{x_1(s_t) + x_2(s_t)}{2})^2}\\
                        % & = \frac{1}{2}\sqrt{x_1(s_t)^2+x_2(s_t)^2+2|x_1(s_t)||x_2(s_t)|\cos \Theta}\\
                        & = \frac{1}{2}\sqrt{2+2\cos \Theta}\\
                        & = \frac{1}{2}\sqrt{4\cos^2 \frac{\Theta}{2}}\quad\because \quad \Theta \in [0^\circ,180^\circ], \quad\therefore\quad \cos \frac{\Theta}{2} \geq 0\\
                        & = \cos \frac{\Theta}{2}\\
        \mu(s_t)  &= \frac{ \mathbf{x}(s_t)}{\mathbf{R}(s_t)}, \quad k(s_t) = \frac{\mathbf{\mathbf{R}(s_t)}(p-\mathbf{R}^2(s_t))}{(1-\mathbf{R}^2(s_t))} \quad\propto \quad\mathbf{R}(s_t)
\end{split}
\end{align}
Hence, when the ensemble number is set to 2, the von Mises-Fisher distribution not only incorporates the angle information but also includes information about the mean direction. Therefore, in this study, we prefer to use the more informative von Mises-Fisher distribution over merely the angle between gradients to quantify the uncertainty of gradient directions.

\section{Additional Experiments}\label{APP:addexp}

Beyond the experiments detailed in the main body of the paper, we have undertaken comprehensive validations to in-depth demonstrate our method's effectiveness. Given the limitations on length in the main text, we've included these additional experimental results in this section. These supplemental experiments encompass:
\begin{enumerate}
\item[(1)] Further experiments on Mujoco (Appendix \ref{SE:add_mujoco})
\item[(2)] An ablation study on the update-to-data ratio (Appendix \ref{SE:addutd})
\item[(3)] Analysis of Hindsight Experience Replay's performance when integrated with vMFER (Appendix \ref{SE:addher})
\end{enumerate}

\subsection{Additional Experiments on Mujoco}\label{SE:add_mujoco}

\begin{figure}[H]
\centering
    \subfloat[InvertPendulum-TD3]{
    \includegraphics[width=.4\columnwidth]
    {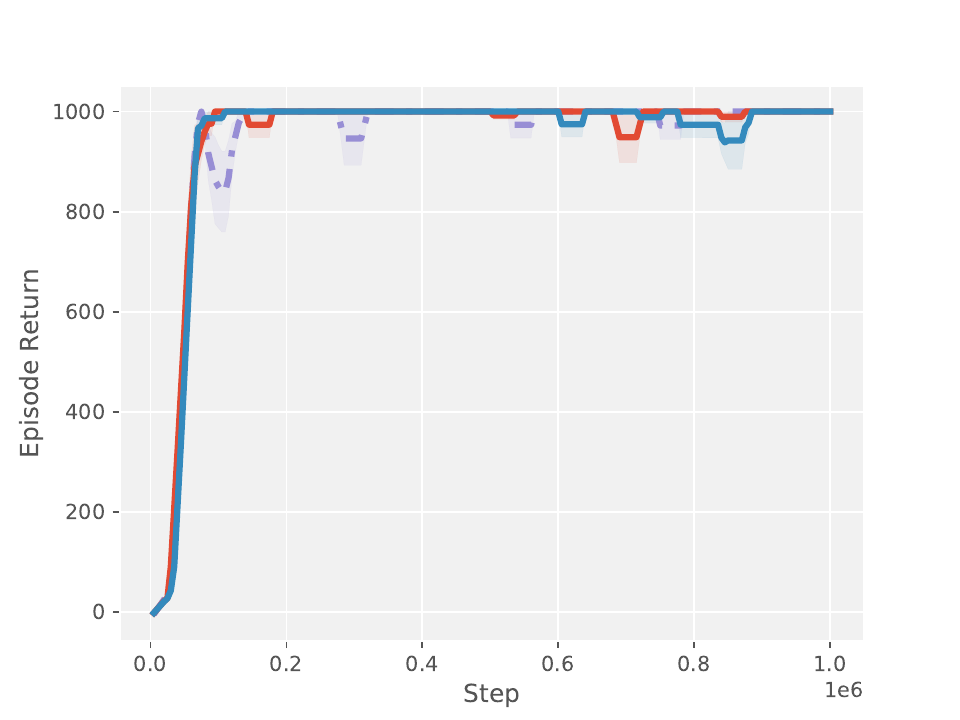}
    \label{Fig:iptd3}
    }
    \subfloat[InvertPendulum-SAC]{
    \includegraphics[width=.4\columnwidth]
    {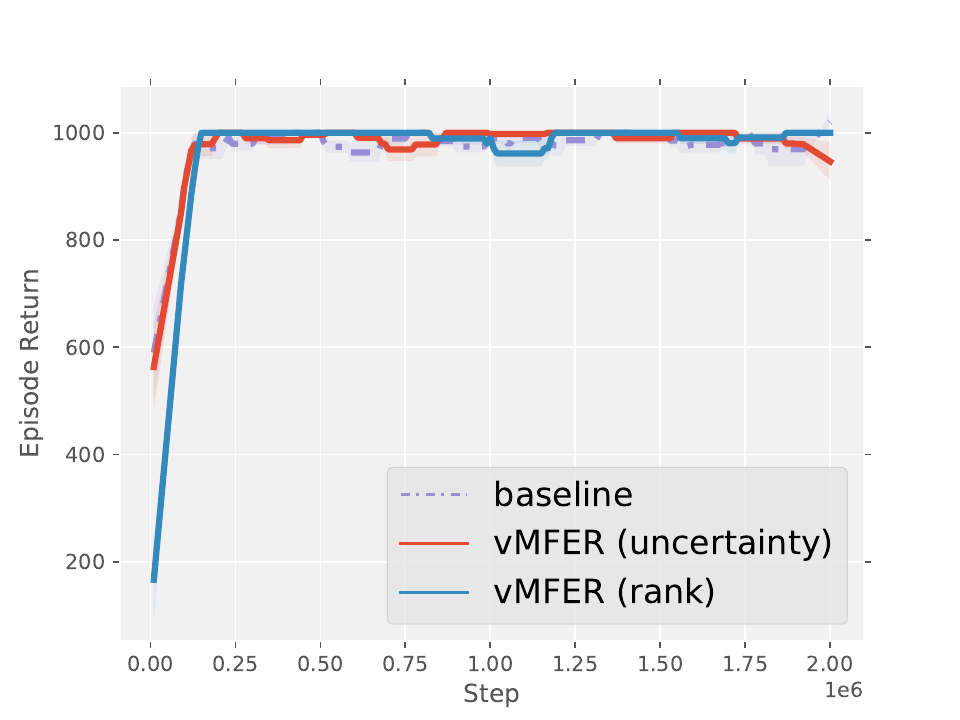}
    \label{Fig:ipsac}
    }\hspace{5pt}
\caption{ Performance in Mujoco robotic control environment -- InvertPendulum  with our algorithm compared to pure TD3 and SAC algorithm}
\label{Fig:mujocoiptd3sac}
\end{figure}

We conducted an additional experiment on the InvertedPendulum to investigate the impact of our vMFER algorithm on the performance of the pure SAC and TD3 algorithms, as depicted in Figure 
{\ref{Fig:mujocoiptd3sac}}.

In the relatively simple Mujoco robotic control environment, such as InvertPendulum, we observed that various methods could converge relatively quickly. However, overall, the policy learned using the vMFER method demonstrated more stable performance compared to those learned without employing vMFER.

\subsection{Ablation Experiment on Update-to-data Ratio}\label{SE:addutd}

\begin{figure}[!htbp]
    \centering
    \begin{minipage}{0.45\columnwidth}
    \subfloat[Ant-UTD1]{
    \includegraphics[width=.5\columnwidth]{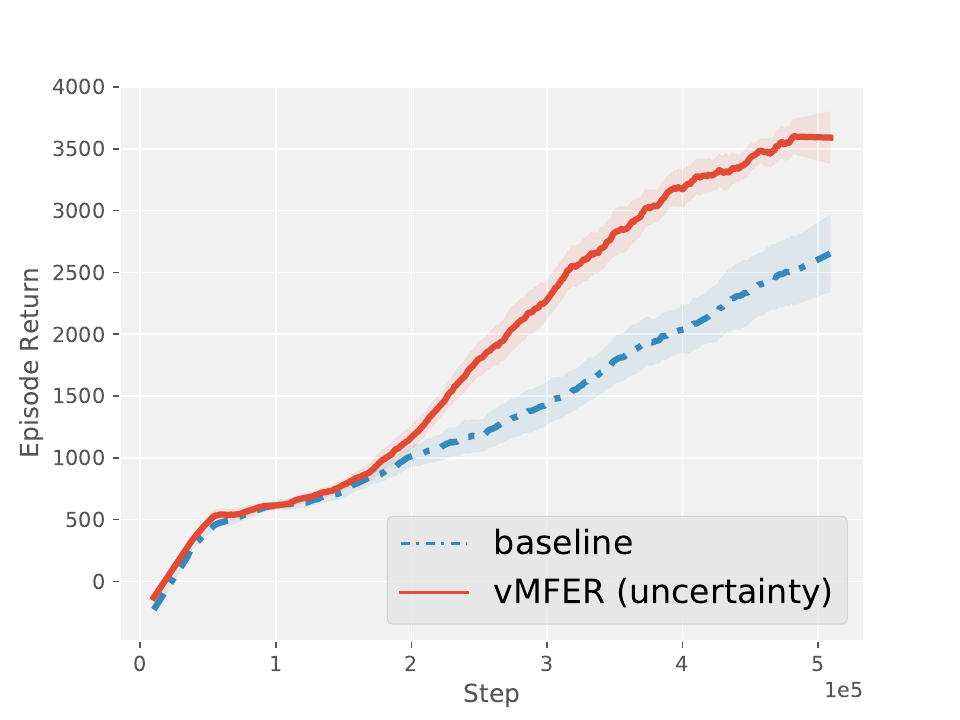}
    \label{Fig:antutd1}
    }\hspace{-15pt}
    \subfloat[Ant-UTD3]{
    \includegraphics[width=.5\columnwidth]{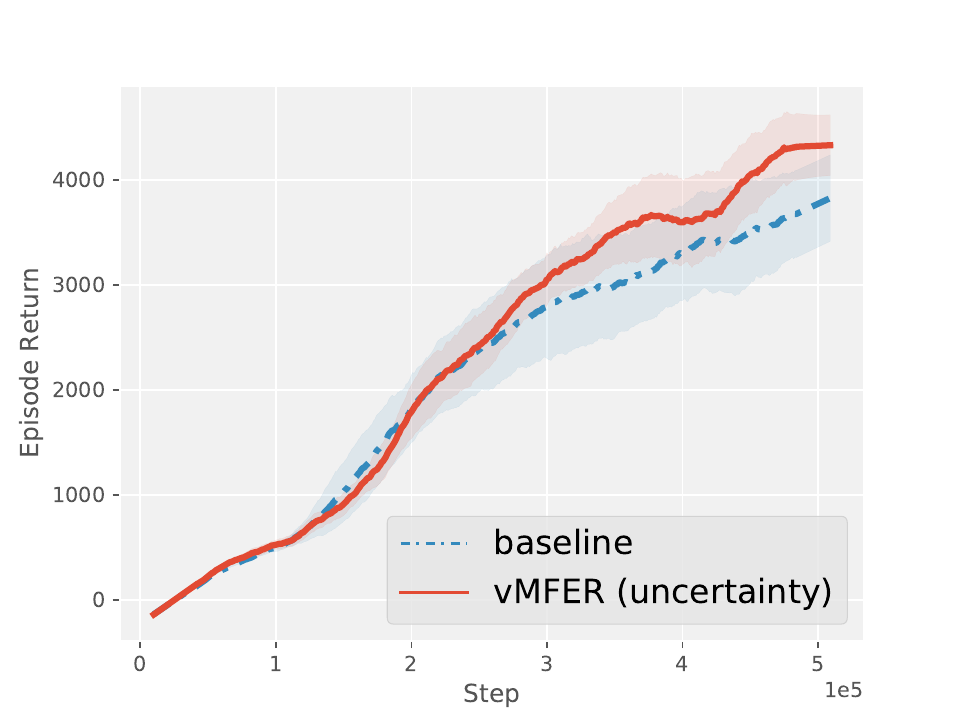}
    \label{Fig:antutd3}
    }\\
    \subfloat[Ant-UTD5]{
    \includegraphics[width=.5\columnwidth]{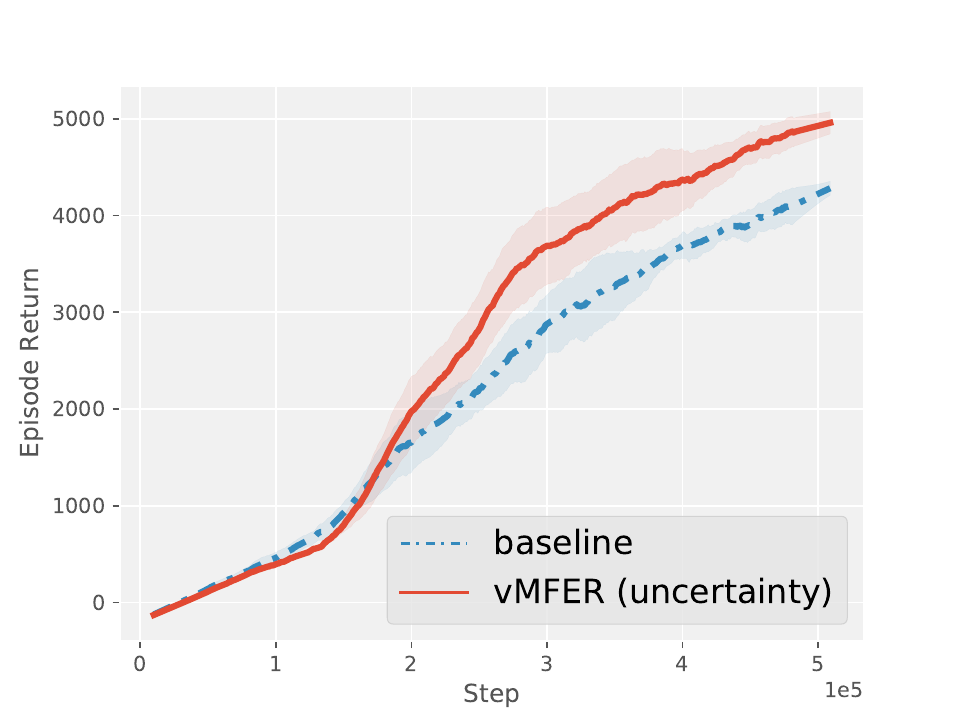}
    \label{Fig:antutd5}
    }
    \hspace{-15pt}
    \subfloat[Uncertainty]{
    \includegraphics[width=.5\columnwidth]{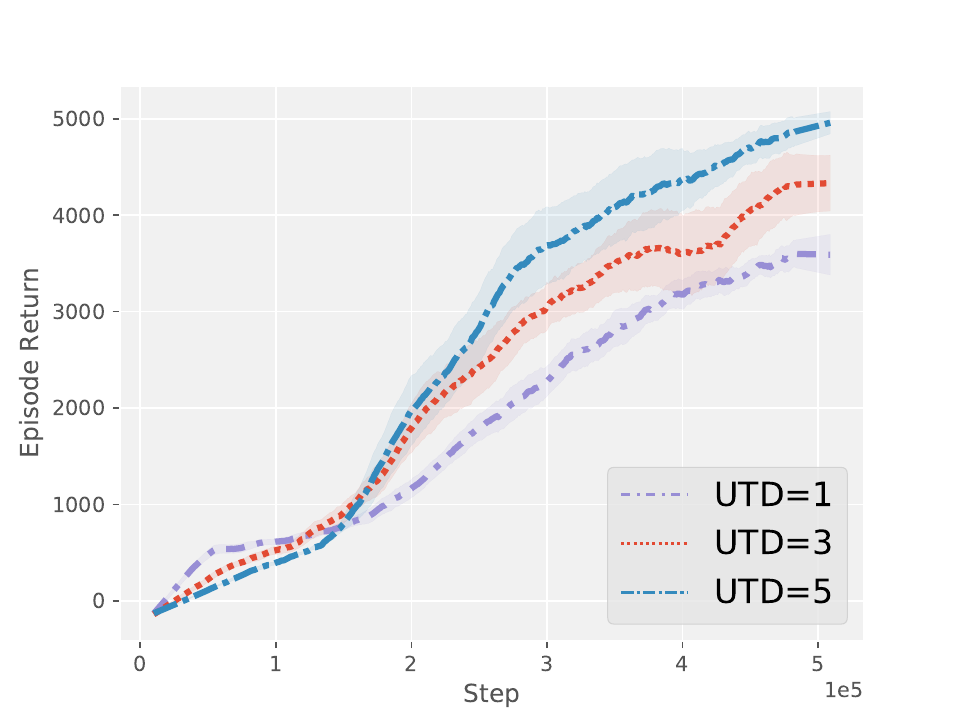}
    \label{Fig:uncertainty_nolegend}
    }
    \end{minipage}
    \hspace{-20pt}
    \begin{minipage}{0.57\columnwidth}
         \subfloat[Summary]{
    \includegraphics[width=1\columnwidth]{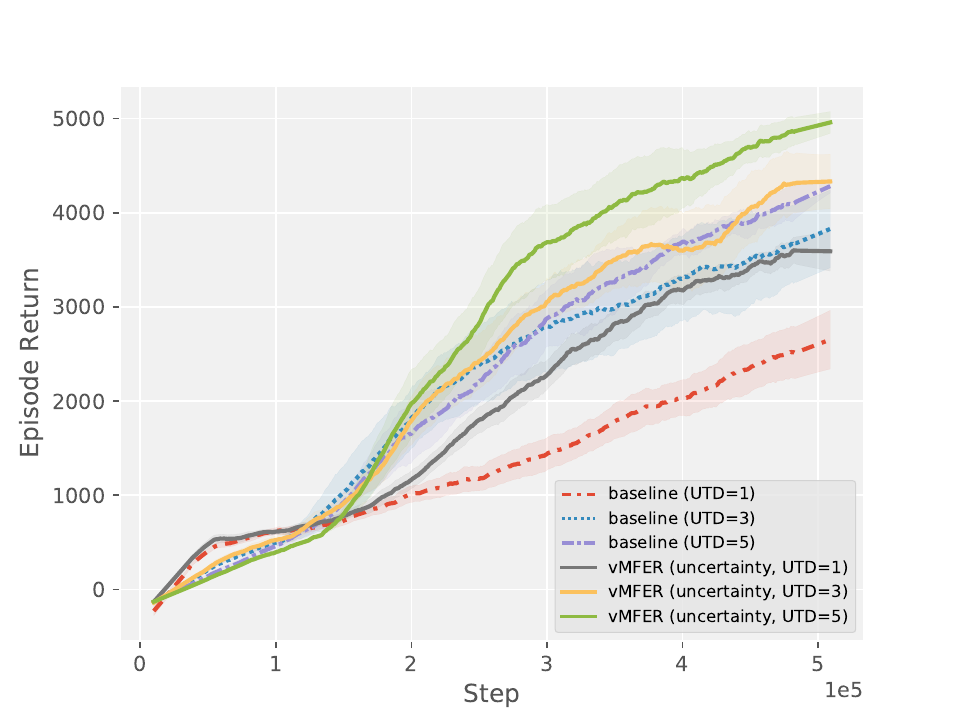}
    \label{Fig:summary}
    }    \hspace{5pt}
    \end{minipage}
\caption{The ablation study on the update-to-data (UTD) ratio}
\label{Fig:ablationofutd}
\end{figure}

We performed an ablation study on the update-to-data (UTD) ratio, and the results obtained after 500K training steps on the Ant environment are presented in Figure {\ref{Fig:ablationofutd}}. Based on these results, we can draw two conclusions.

Firstly, the addition of UTD does lead to performance gains and improved training efficiency, as demonstrated by both our algorithm and the baseline algorithm.

Secondly, the performance of our algorithm at UTD=1 is comparable to the performance of the baseline algorithm at UTD=3. This suggests that the baseline algorithm with UTD=3 requires twice the amount of training effort on the critic to achieve similar results as our algorithm at UTD=1. In contrast, our algorithm does not increase the cost of training the critic compared to the baseline algorithm. Instead, we utilize a data sampling approach to improve training efficiency while maintaining a lower resource consumption level.

\subsection{Combine Hindsight Experience Replay with vMFER}\label{SE:addher}
\subsubsection{Hindsight Experience Replay}%Introduction of 
One challenge in RL is that the agent may not receive any immediate reward for its actions, making it difficult to learn from past experiences. Hindsight Experience Replay (HER)  is an algorithm that addresses this problem by reframing past experiences in terms of their outcomes \cite{andrychowicz2017hindsight}.
HER saves all past experiences in a replay buffer and modifies the original goals with the achieved goals during the sampling process. Additionally, the reward function of past experiences is also modified to reflect the achieved goal rather than the original goal. This approach allows the agent to learn more from experiences and alleviate the sparse rewards problem.
HER has been proven to be effective in solving sparse rewards problems where the agent only receives a non-zero reward at the end of the episode. It has shown success in tasks such as robotic manipulation and navigation.

\subsubsection{Combining vMFER with Hindsight Experience Replay}\label{SE:vmfer+HER}

\begin{figure}[H]
    \centering
    \hspace{-25mm}
    \subfloat[FetchPush-v1]{
    \includegraphics[width=.38\columnwidth]{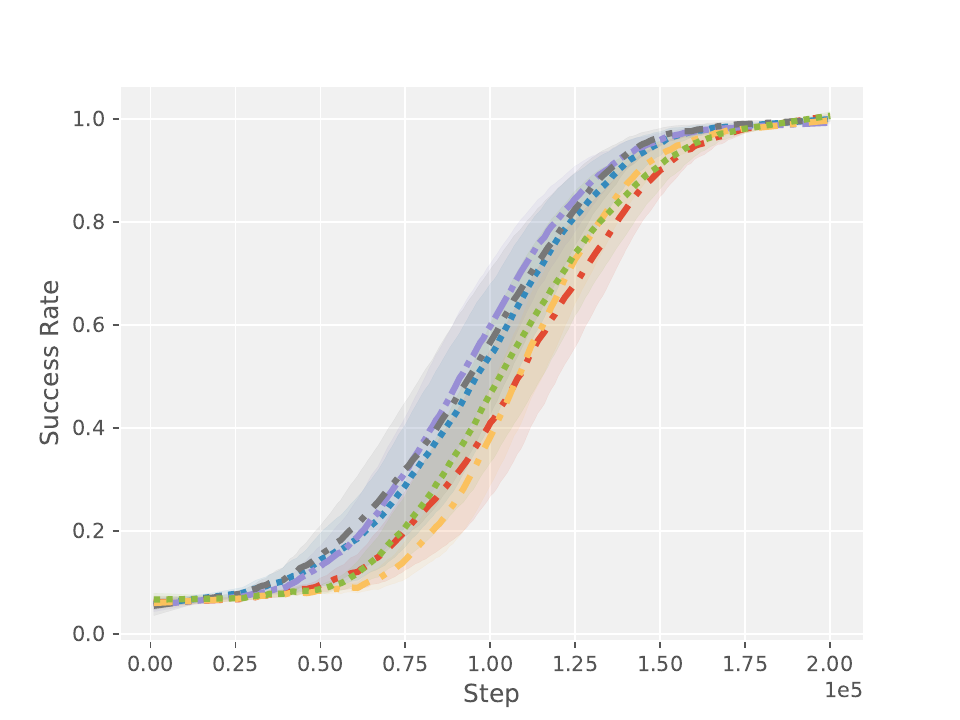}
    \label{Fig:env-Hopper}
    }
    \hspace{-5mm}
    \subfloat[FetchPickAndPlace-v1]{
    \includegraphics[width=.38\columnwidth,]{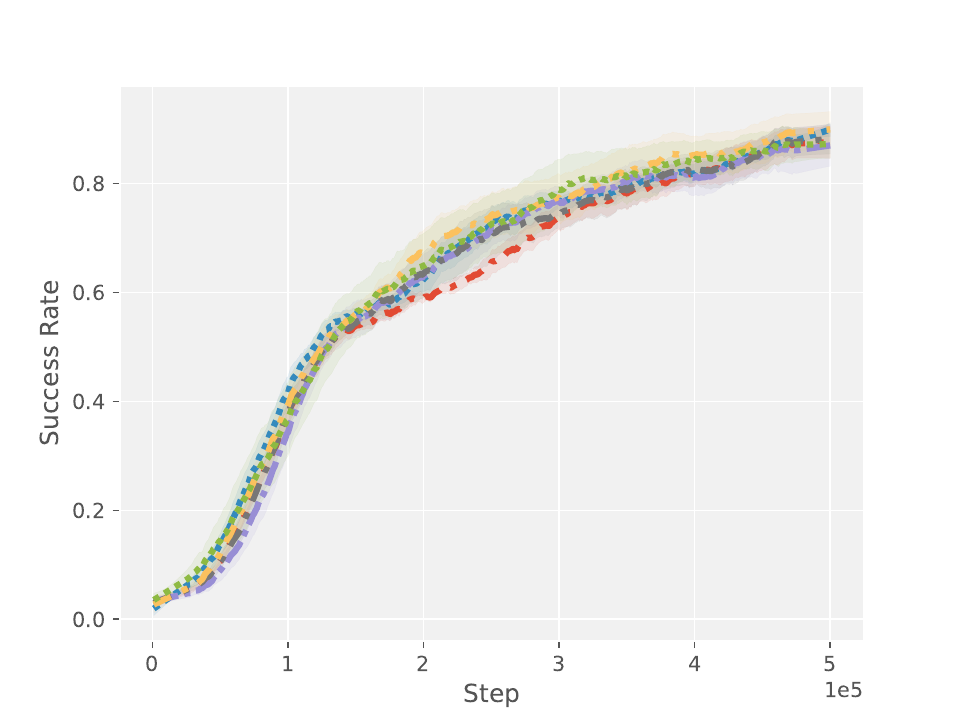}
    \label{Fig:env-Ant}
    }
    \hspace{-5mm}
    \subfloat[FetchSlide-v1]{
    \includegraphics[width=.38\columnwidth,]{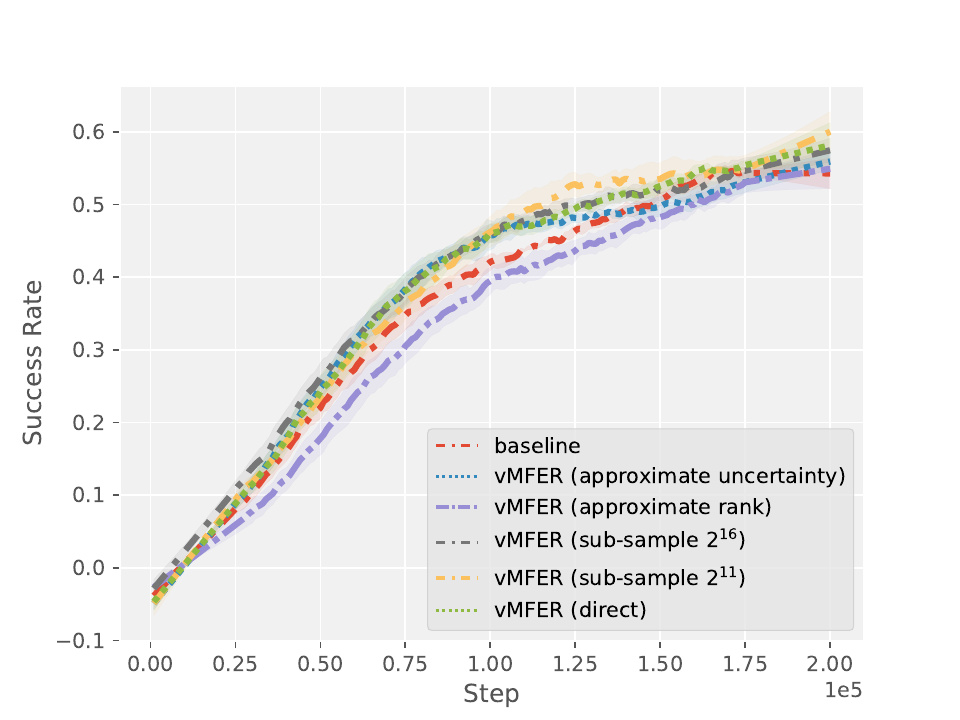}
    \label{Fig:env-HalfCheetah}
    }\hspace{-25mm}
    % \hspace{5pt} ,height=2.5cm
    \caption{Evaluating the impact of our vMFER algorithm on HER+TD3 in sparse rewards robotic arm control tasks, the parameter of the sub-sample method represents the sub-sample size ($2^{11}$,$2^{16}$). }
    \label{Fig:HER_TD3_Fetch}
\end{figure}
Hindsight Experience Replay (HER)  is a widely used algorithm for solving problems with sparse rewards. 
As dense rewards are not commonly available in complex tasks \cite{plappert2018multi}, we opt for a sparse reward setting and combine vMFER with HER \cite{andrychowicz2017hindsight} to assess their effectiveness.

Our algorithm utilizes separate transition labeling to facilitate sampling during actor updates.
HER itself can be considered a form of data augmentation, which enhances a transition $(s,a,g)$ to multiple transitions $(s,a,g'), g'\in G$ where the potential goal $G$ is chosen by HER and $g$ represents the goal of the task.
The buffer with additional transitions will increase the number of transitions available for training. However, if we continue to utilize the original vMFER configuration, which computes uncertainty values for mini-batch transitions at once, this may result in an excessive lag in buffer uncertainty, causing the algorithm to deviate from our expected performance, compared to the outperformance in Mujoco environments.

To address this problem brought by the explosion of augmented data, we attempted three approaches. The first involves approximating the update of $p(s_j,g)$, using the formula $p(s_j,g)\xleftarrow{}0.9 \times p(s_j,g)+ 0.1 \times (\exp(\mathbf(R(s_j,g)\mu(s_j,g)^{\textbf{T}}x(s_j,g)))$, which doesn't consider the impact of data augmentation brought by HER. The second approach, known as the sub-sample method, involved uniformly sampling a large batch of transitions from the replay buffer for ranking and sampling, instead of sampling from the entire augmented buffer directly. The last can be referred to as the direct method, which focuses on directly calculating uncertainty for all the augmented transitions $(s,a,g')$.

\subsubsection{Performance of vMFER Combined with HER in Robotoic Control with Sparse Rewards}\label{SE:Fetchexp}

We utilize the vMFER method for robotic learning in complex control environments with sparse rewards. Our baseline algorithm is  HER+TD3. Then, we compare the approaches mentioned in Section \ref{SE:vmfer+HER} to the pure HER+TD3 algorithm.
The results, as shown in Figure {\ref{Fig:HER_TD3_Fetch}}, demonstrate that vMFER outperforms the baseline algorithm, with a notable advantage in the FetchPush task. 
While the advantage was less prominent in the FetchSlide task, vMFER's performance is still comparable to the baseline algorithm.

\begin{figure}[!hbpt]

    \centering
    \includegraphics[width=.45\columnwidth]{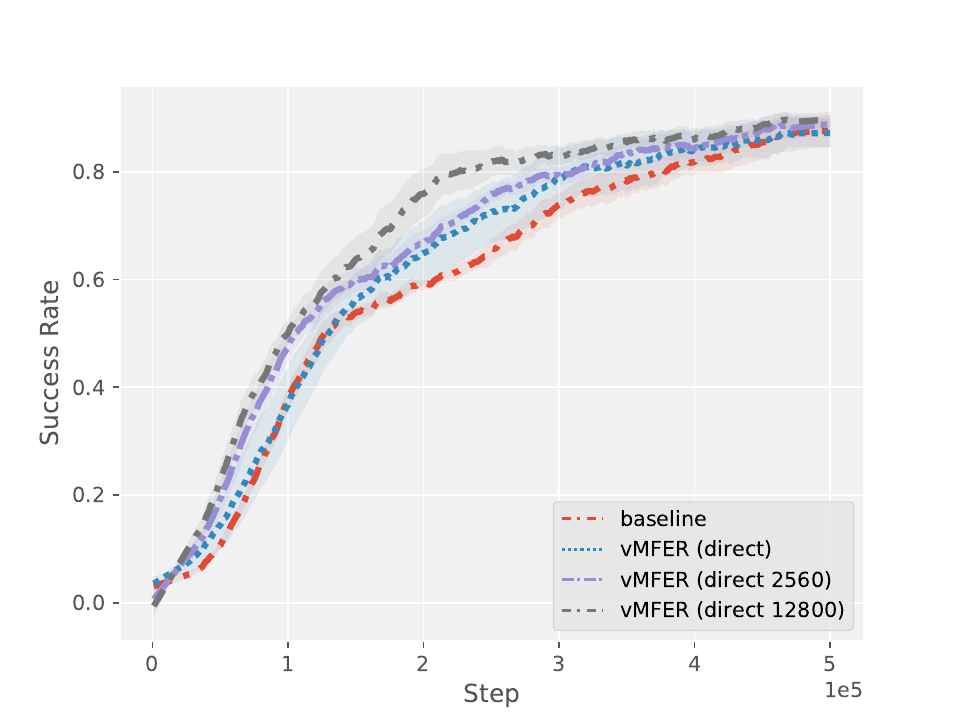}
    \caption{The impact of varying the number of transitions used for calculating the uncertainty once updated.}
    \label{Fig:ablationOfFetchDirect}
    \setlength{\abovecaptionskip}{0pt}
\setlength{\belowcaptionskip}{-20pt}
\end{figure}

We carry out additional experiments using the direct method and compare the effects of using varying numbers of transitions (256, 2560, and 12800) for calculating uncertainty once updated.
The results show that increasing the number of used transitions improves our algorithm's performance,  shown in Figure {\ref{Fig:ablationOfFetchDirect}}.

\section{Hyper-parameter}\label{APP:hyperp}
The table below shows the hyper-parameters for the algorithms used in our experiments.
\begin{table}[h!]
  \begin{center}
    % \caption{Your first table.}
    \scalebox{0.8}{
    \begin{tabular}{l|c|c|c|c} 
    \hline
      \textbf{Hyper-parameter} & \textbf{TD3} & \textbf{SAC}& \textbf{HER+TD3}& \textbf{PER+SAC}\\
      \hline
      Critic Learning Rate & $10^{-3}$ & $3\cdot 10{-4}$ &  $10^{-3}$ & $3\cdot 10{-4}$\\
      Actor Learning Rate & $10^{-3}$ &  $3\cdot 10{-4}$ &  $10^{-3}$ &  $3\cdot 10{-4}$\\
      Optimizer & Adam & Adam & Adam & Adam\\
      Target Update Rate ($\tau$)  & $5\cdot10{-3}$& $5\cdot10{-3}$& $5\cdot10{-3}$& $5\cdot10{-3}$\\
      Batch size & 256 & 256 & 256 & 256\\
      Iterations per time step & 1 & 1 & 1 &1 \\
      Discount Factor ($\gamma$) & 0.99 & 0.99 & 0.99 &0.99\\
      Normalized Observations & False& False & True &False\\
      Exploration Policy & $N(0,0.1)$& None & $N(0,0.1)$ & None\\
      Number of hidden units per layer &256&256&256 &256\\
      Number of hidden layers &2&2&3 &2\\
      Buffer size &1e6&1e6&3e6&1
      e6\\
      Nonlinearity&Relu&Relu&Relu&Relu\\
      Target Entropy  ($\bar{\mathcal{H}}$)& None &-dim($\mathcal{A}$)&None&-dim($\mathcal{A}$)\\
      Sampling method for policy evaluation&Uniform&Uniform&Uniform&Rank(PER)\\
      \hline
    \end{tabular}
    }
  \end{center}
\end{table}

\section{Discussion - Why Direction, Not Magnitude}
Erroneous/divergent/non-convergent gradient direction and magnitude both impact the learning process. Comparatively, an incorrect gradient descent direction is less acceptable than an incorrect magnitude, as the former implies invalid optimization of the actual objective, while the latter still optimizes the objective. 
Therefore the emphasis of this paper is on gradient direction. 
Nevertheless, in the future, we will explore methods that can simultaneously address both gradient direction and magnitude.

\section{Shooting Environment}
In the toy environment detailed in Section \ref{SE:NecOFun}, players take on the role of a shooter. Following each shot, a reward is given based on the landing position of the shot, after which the game is reset.
\paragraph{State.}The initial state for every attempt is identical, set at $s_0 = (-0.5, -0.5)$. 
\paragraph{Action.}The action space is defined as ${ a \in \mathrm{R}^2}$. An optimal action, $a^* = {(-0.5, -0.5)}$, is predetermined to guarantee a hit at the center of the target. 
\paragraph{Reward.} The reward function is defined as $r=-||a-a^*||_2$.
\paragraph{Transition Function.} 
Given that this scenario constitutes a one-step MDP, the environment resets after each shot. Therefore, there is no necessity to define a state transition function or a state space.

\section{Computational Complexity}

We acknowledge the importance of computational cost considerations. To address this, we have optimized our method to ensure that its time cost exceeds the baseline by only $10\%$ when using a GPU. Our primary goal is the timely update of transition uncertainty for each sampled batch size. This is achieved by computing $\frac{\partial l_i(s_t,a)}{\partial a}$, which has dimensions of \textbf{ensemble\_num$*$batch\_size$*$action\_dim}.

For the calculation of the loss action derivative, we utilize PyTorch's built-in function. While this incurs some additional computational load, it is justified by the benefits our method offers. For further clarity, we provide the logic of our code below for your reference.
{\small{
\begin{verbatim}
# losses (ensemble_size*batch_size*1)
# action (batch_size*action_dim)
for i in range(ensemble_num): # ensemble_size=2 in SAC and TD3
    grad_temp = torch.autograd.grad(
                    losses[i,::].sum(),
                    action,retain_graph=True)[0] 
    # grad_temp (batch_size*action_dim)
    grad +=[grad_temp.unsqueeze(0)]	
grad = torch.cat(grad,dim=0) # ensemble_num*batch_size*action_dim
\end{verbatim}
}}

On a 2080ti GPU machine, SAC algorithm updates averaged {\color{blue}0.0208} s. When combining with vMFER (uncertainty), this rose to {\color{blue}0.0215} s, and vMFER (rank) updates averaged {\color{blue}0.0228} s, covering policy evaluation, policy improvement, and vMFER-required uncertainty updates. 

\section{Algorithm Combined with vMFER}\label{APP:algo}
\begin{algorithm}[!hbtp]
\caption{Von Mises-Fisher Experience Resampling (Based on SAC)}
\label{Algo:vMFER_sac}
\footnotesize 
\begin{algorithmic}[1]
\State
    Initialize replay buffer $\mathcal{D}$ , ensemble number $N$ and target entropy $\bar{\mathcal{H}}$
\State  
    Initialize critic networks $\{Q_{\phi_i}\;|\; i\in[1,N]\}$ and actor network $\pi_\theta$ with random parameters $\{{\phi_i}\;|\; i\in[1,N]\}$, $\theta$.
\State  Initialize  temperature hyperparameter $\alpha$
\State
    Initialize target network $\{{\phi_i}{'} \leftarrow {\phi_i}\;|\; i\in[1,N]\}$, $\theta{'} \leftarrow \theta$
\State
    Initialize sampling factors $p_j = 1$ for each transition of $\mathcal{D}$
\For{$t$=1 to $T$ }
    \State sample action  $a\sim\pi_{\theta}(s)$
    \State store transition $(s,a,r,s')$ in $\mathcal{D}$ 
    \State sample mini-batch of $b$ transitions $(s,a,r,s')$ from $\mathcal{D}$
    \State $y\leftarrow r+ \gamma \left(\min_{i=1,2}Q_{\phi_i}(s',a') -\alpha \log(\pi_\theta(a'|s'))\right),\;\; a'\sim \pi_\theta(s')$
    \State  $\phi_i \leftarrow \mathop{argmin}_{\phi_i} b^{-1}\sum (y - Q_{\phi_i}(s,a))^2$ \Comment{Update critics}
    \For{$j = 1$ to $b$}
    \State  $(s_j,a_j,r_j,s'_j)\sim P(j) = \frac{p_j}{\sum_m p_m}$  \Comment{ Eq. ({\ref{Eq:Baye}}), {\color{brown}  Resample transition} }
    \State$\hat{a}_j\leftarrow \pi_{\theta}(s_j)$ \Comment{Sample action }
     \State   $l_i(s_j,\hat{a}_j) =\alpha \log(\pi_\theta(\hat{a}_j|s_j)) -Q_{\phi_{i} }(s_j,\hat{a}_j),\;  \;i\in [1,N] $ \Comment{{\color{brown}  Calculate the actor losses}}
    \State $ g_i = \frac{\partial l_i(s_j,a)}{\partial a} |_{a=\hat{a}_j},\;  \;i\in [1,N] $ \Comment{{\color{brown}  Calculate gradients of losses}}
    \State $p_j\leftarrow$ {\color{brown} Update Sampling Factor $p_j$ (Algorithm \ref{Algo:usfpj})}
    \EndFor
    \State $\nabla_\theta J(\theta) =-b^{-1}\sum_j \nabla_\theta \max\limits_{i\in[1,N]} l_i(s_j)$ \Comment{Update actor}
    \State $\nabla_\alpha J(\alpha) = -b^{-1}\nabla_\alpha\sum_j -\alpha\log\pi_\theta(\hat{a}_j|s_j)-\alpha \Bar{\mathcal{H}}$ \Comment{Update temperature hyperparameter}
    \State$\phi_i'\leftarrow \tau \phi_i+(1-\tau)\phi_i'$, $\theta'\leftarrow\tau\theta+(1-\tau)\theta'$ \Comment{Update target networks}
\EndFor
\end{algorithmic}
\end{algorithm}
In Algorithm \ref{Algo:vMFER_sac}, we integrate vMFER with SAC. The sections marked in brown indicate the aspects where our version differs from the original SAC.

\end{document}